\documentclass[lettersize,journal]{IEEEtran}

\usepackage{cite}
\usepackage{amsmath,amsfonts}
\usepackage{algorithmic}
\usepackage{array}
\usepackage[caption=false,font=normalsize,labelfont=sf,textfont=sf]{subfig}
\usepackage{textcomp}
\usepackage{stfloats}
\usepackage{url}
\usepackage{verbatim}
\usepackage{graphicx}
\usepackage{color}
\usepackage{bm}
\usepackage{multirow}

\newcommand{\black}					{\color{black}}

\usepackage{tabularx} 

\def\BibTeX{{\rm B\kern-.05em{\sc i\kern-.025em b}\kern-.08em
    T\kern-.1667em\lower.7ex\hbox{E}\kern-.125emX}}
\usepackage{balance}
\begin{document}
\title{On the \black Optimized \black Use of Non-Orthonormality Constraints \black for \black 
the 
\black Quasi-Static \black 
\black INS 
Alignment of 
\black Autonomous Underwater and Surface Vehicles}

\author{Carlos Renato C.~Durão,
Felipe O.~Silva,~\IEEEmembership{Senior Member,~IEEE},
Itzik Klein,~\IEEEmembership{Senior Member,~IEEE},
Vinícius M.~G.~B.~Cavalcanti,
Adriano Frutuoso,
Ettore A.~de~Barros,
and Jay A.~Farrell,~\IEEEmembership{Fellow,~IEEE}

\thanks{
Carlos Renato C. Durão and Felipe O. Silva are with the
Department of Automatics,
Federal University of Lavras,
Lavras, MG, Brazil
(e-mail:
renatodurao1963@gmail.com;
felipe.oliveira@ufla.br).
}
\thanks{
Itzik Klein is with the
Hatter Department of Marine Technologies,
University of Haifa,
Haifa, Israel
(e-mail: kitzik@univ.haifa.ac.il).
}
\thanks{
Vinícius M. G. B. Cavalcanti is with the
Department of Electrotechnics and Automation,
Fluminense Federal Institute of Education, Science and Technology,
Itaboraí, RJ, Brazil
(e-mail: vinicius.cavalcanti@gsuite.iff.edu.br).
}
\thanks{
Adriano Frutuoso is with the
Department of Higher Education,
Federal Institute of Education, Science and Technology of Amazonas,
Manaus, AM, Brazil
(e-mail: adriano.frutuoso@ifam.edu.br).
}
\thanks{
Ettore A. de Barros is with the
Unmanned Vehicles Laboratory,
University of São Paulo,
São Paulo, SP, Brazil
(e-mail: eabarros@usp.br).
}
\thanks{
Jay A. Farrell is with the
Calibration, Localization and Mapping Team,
Zoox, Inc.,
Foster City, CA, USA
(e-mail: jafarrell@gmail.com).
}

\thanks{This work was financed in part by the Minas Gerais State Agency for Research and Development (FAPEMIG), under grant APQ-04659-22, in part by the Brazilian National Council for Scientific and Technological Development (CNPq), under grant 312194/2022-6, 
in part by the Research Development Foundation (FUNDEP - MOVER), under grant 27192.02.02/2021.01.00, \black and in part by the Federal University of Lavras (UFLA), under grant 23090.000874/2026-48\black.}}

\markboth{Journal of \LaTeX\ Class Files,~Vol.~18, No.~9, September~2020}%
{How to Use the IEEEtran \LaTeX \ Templates}

\maketitle

\begin{abstract}
Inertial navigation systems 
\black are specialized navigation apparatuses that equip almost all autonomous underwater and surface vehicles. 
They \black require precise initial alignment, \black i.e., determination of their initial attitude, \black which is typically achieved: \black (a) in quasi-static conditions (whenever possible); and (b) \black in two stages: Coarse Alignment (CA), 
using methods like TRI-axis Attitude Determination (TRIAD), and Fine Alignment (FA), 
via \black Zero Velocity Update (ZVU)-\black
based Extended Kalman Filtering (EKF). 
\black However, conventional 
methods suffer from slow convergence and limited bias 
\black estimability\black. In response, this paper 
\black introduces: \black (a) an optimized version of a recently proposed CA method, namely, TRIAD with \black Coarse Bias Estimation (TRIAD-CBE); \black
and (b) \black a novel FA EKF observation model that incorporates \black Non-Orthonormality (NON) \black 
error constraints derived from 
TRIAD, 
directly linking these errors to the inertial sensor biases. 
\black As validated through extensive Monte Carlo \black (MC) \black 
simulations, as well as real-world experiments using two \black Inertial Measurement Units (IMUs) \black 
of different grades, our approaches substantially accelerate the convergence of misalignment and bias estimates (from minutes to seconds), while maintaining accuracy/precision comparable to traditional techniques. 
\end{abstract}



\section{INTRODUCTION} \label{sec:intro}

\black Autonomous Underwater and Surface Vehicles (AUVs/ ASVs) play a critical role in maritime scientific research and industrial operations, driving advances in oceanography, climate studies, marine biology, seafloor mapping, and offshore resource exploration \cite{ZHANG2023113861}. For AUVs, navigation and localization are more challenging than for ASVs because Global Navigation Satellite System (GNSS) signals attenuate rapidly underwater \cite{paull14}. Alternative sensors such as Doppler velocity logs, 
sonars, depth meters, and flow speed sensors can help mitigate the absence of GNSS~\cite{pinto22,meurer20,fallon13,song18}. Nevertheless, most AUV/ASV navigation solutions ultimately rely on systems that do not depend on external signals or environmental feature recognition—most notably, the Inertial Navigation System (INS). A key limitation of INS technology is its ``unbounded" navigation error growth, which is directly tied to the quality of its inertial sensors. Furthermore, initializing an INS, particularly its attitude—a process known as alignment—can be cumbersome. Ideally, an INS should align as accurately, quickly, and autonomously as possible \cite{silva16}. This is especially feasible when the AUV/ASV is equipped with intermediate-grade or higher inertial sensors and alignment is performed under quasi-static conditions, such as on a pier, the deck of a large vessel, or a moored ship \cite{mcphail09,sarda19}.\black

In general, the \black quasi-static alignment of INSs \black 
is carried out in two steps: \black C\black oarse \black A\black lignment~(CA) and \black F\black ine \black A\black lignment~(FA)~\cite{jekeli00}. In the CA, a first \black (analytical and non-optimal) \black estimate of the 
\black rotation \black matrix is obtained from the gravity and Earth rate vectors, measured by \black \black quasi-static \black 
accelerometers and gyroscopes \cite{S2000}. 
Several formulations for CA have been proposed in the literature. \black The 
\black TRI-axial Attitude Determination
\black (TRIAD) \black method \cite{markley14} is the basis for 
\black many \black of them, \black wherein an orthogonal basis of unity vectors observed in both frames of interest is used to analytically compute the INS initial attitude. \black 
A variation of 
TRIAD 
was presented in \cite{jiang1998}, where non-normalized versions of the \black observation vectors \black were adopted. \black Silva \textit{et al.}~\cite{silva18TIM} expanded the idea by using the latter vectors to construct a non-orthogonal triad of observations, in order \black to \black purposely \black preserve 
the 
Non-Orthonormality~(NON) errors \black of 
\black the estimated 
\black rotation \black matrix. In the same paper, \black the relationship \black between the inertial sensors biases and the NON errors was analytically determined, which allowed for the \black C\black oarse \black B\black ias \black E\black stimation (CBE) \black during the CA \black(reason why Silva \textit{et al.} \cite{silva18TIM} called 
this method 
TRIAD-CBE). 

Due to the 
\black oversimplified assumptions the CA is based on, its rotation matrix estimate is \black generally \black non-optimal. Therefore, aiming \black to improve \black the latter, 
\black by \black 
also estimating the inertial sensor biases, the FA is executed after the CA~\cite{titterton04}\black. Most of the FA procedures proposed in the literature are based on 
\black an algorithm that \black uses the 
INS equations and a \black linearized \black error propagation model within an \black \black E\black xtended Kalman \black F\black ilter~(EKF) \black structure to estimate the initial platform misalignment and sensors biases\black \cite{bar-itzhack88}. \black 
In a traditional \black quasi-static \black FA implementation, \black also referred to as \black Z\black ero \black V\black elocity \black U\black pdate (ZVU)-based approach, \black the Earth-referenced velocity error vector is used as measurement\black\cite{groves13,engelsman2023information}\black. In \black such an \black implementation, some components of the sensors biases are weakly estimated\cite{silva17} 
or require a long time to converge. 
\black This is why, in general, the benefits \black that \black the traditional FA brings to the accuracy and alignment time of the previously CA-estimated \black attitude \black 
are too modest.

\black In an attempt to accelerate the convergence rate of the FA, Silva \textit{et al.} \cite{silva14,silva18MEAS} proposed to augment the ZVU-based EKF observations, with \black C\black onstant Earth \black R\black ate and \black G\black ravity \black U\black pdates~(CERGUs). \black Despite the method's effectiveness, the authors did not assess the possible negative impacts of the correlated process and measurement noise originated from CERGU observation. 
Also aiming \black to improve \black 
accuracy and convergence time, \black Lin, Miao and Zhou 
\cite{RefB8} 
proposed \black an initial alignment method for 
INSs based on the backtracking process, which reuses short-time measurement data 
into \black both \black CA and FA. \black 
\black A similar approach—a Backtracking Kalman Filter (BKF)—was proposed in \cite{RefB6} to address similar challenges. 
This method, however, was not specifically developed for strapdown INSs, 
but rather for rotary INSs. 

\black Differently 
from recursive methods (like 
BKF), \black a novel \black FA \black approach \black \black for 
INSs \black has been proposed in \cite{RefB7}. \black 
\black Such a \black methodology expedites bias error calculations by utilizing quaternion-based analytical relationships, which bypasses the slow convergence behavior associated with recursive algorithms. The proposed approach 
\black demonstrated \black comparable accuracy to traditional \black FA \black 
methods within 20 
\black seconds\black. To improve the 
INS FA when low-cost sensors are used, 
a measurement strategy combining different 
INS aiding types 
\black was also proposed (and compared with five types of adaptive KFs) in 
\cite{RefB9}. 


\black Given the aforementioned context, in \black 
this paper, 
\black we propose: (a) an \black optimized \black 
version of 
TRIAD-CBE, 
\black which incorporates \black 
a \black Weighted \black Least Squares \black (WLS) solution for the \black 
bias 
components \black estimation (hereinafter referred to as OPT-TRIAD-CBE); and (b) \black a new observation model for the \black EKF-based \black FA, 
which \black relies \black 
on the NON error equations originally derived for use in TRIAD-CBE. Given the optimality of EKF w.r.t. \black the original \black TRIAD-CBE (a pure analytical procedure), the purpose is to evaluate how the 
accuracy and alignment time \black of FA \black can benefit from the latter model, \black as well as how it compares with the newly proposed OPT-TRIAD-CBE \black. As the main contributions 
\black of this paper, we show that: (a) although the proposed model largely improves the FA convergence rate w.r.t. the traditional ZVU-based EKF, the ultimate alignment accuracy obtained from its use does not improve over the one obtained via CA; and (b) 
\black OPT-\black TRIAD-CBE\black-based CA \black not only provides more accurate/precise estimates for the \black estimable \black 
INS biases, as faster as well (especially when the gyroscope noises are high and time-correlated). 

The rest of this paper is organized as follows: The formulation of the traditional TRIAD-based CA and ZVU-based FA methods are described in Section \ref{sec:trad}. TRIAD-CBE, instead, is reviewed in Section \ref{sec:cbe}, \black and its optimized version \black (OPT-TRIAD-CBE) \black presented in Section \ref{sec:cbe-opt}. 
Section \ref{sec:prop} describes the adaptation of TRIAD-derived NON errors for use in an optimal EKF-based framework. Sections \ref{sec:res} 
and \ref{sec:exp}, \black in sequence, \black provide results/assessments of simulated 
tests, 
including a thoughtful Monte Carlo~\black (MC) \black analysis, \black as well as real-world experiments using a high-end turntable and an ASV mounted on a pier. \black In Section \ref{sec:con}, lastly, the conclusions 
are reported.



\section{The Traditional Alignment Method} \label{sec:trad}

As mentioned in Section \ref{sec:intro}, the \black quasi-static \black alignment of \black INSs \black 
is executed in two steps: CA and FA. One of the most commonly used methods \black for \black 
CA is 
TRIAD. In this method, the \black rotation \black matrix \(\textbf{C}_b^n\) from the \black $b$\black-frame \black(body) \black to the \black $n$\black-frame (\black North-East-Down (NED)\black) is determined from the gravity \((\textbf{g})\) and Earth rate \((\textbf{w}_{ie} )\) vectors \black observed \black 
in these two 
frames. 
In the \black $n$\black-frame, we have\cite{groves13}: \black
$\textbf{g}^n= \begin{bmatrix} 
        0 & 0 &g_D\\
\end{bmatrix}^T$,
$\textbf{w}_{ie}^n= \begin{bmatrix}
w_e \cos(L) & 0 & -w_e \sin(L)
\end{bmatrix}^T$, \black
where \(g_D\) and \(w_e\) \black are \black 
the 
gravity and Earth rate \black magnitudes, \black and \(L\) is the 
latitude.

The same vectors can be \black observed \black 
in 
\black $b$\black-frame as \cite{groves13}: \black
$    \textbf{g}^b\approx-\textbf{f}^b=\begin{bmatrix}
        -f_x & -f_y & -f_z\\
    \end{bmatrix}^T$,
$    \textbf{w}_{ie}^b\approx\textbf{w}_{ib}^b=\begin{bmatrix}
        w_x & w_y & w_z
    \end{bmatrix}^T$, \black
where \(\textbf{f}\) and \(\textbf{w}_{ib}\) are the specific force and angular rate vectors measured by \black quasi-static \black 
\black accelerometers and gyroscopes, \black respectively\black. According \black to \black Jekeli \cite{jekeli00}\black, an initial estimate \black(denoted with 
$\hat{}$\;) \black of the \black rotation \black matrix 
can be obtained \black as: \black 
\begin{equation}
    \hat{\textbf{C}}_b^n=\begin{bmatrix}
    (\textbf{g}^n)^T\\
    (\textbf{w}_{ie}^n)^T\\
    (\textbf{g}^n\times\textbf{w}_{ie}^n)^T\\
    \end{bmatrix}^{-1} \begin{bmatrix}
        (\textbf{g}^b)^T\\
        (\textbf{w}_{ie}^b)^T\\
        (\textbf{g}^b\times\textbf{w}_{ie}^b)^T\\
    \end{bmatrix}.
\end{equation}

The \black traditional \black TRIAD-based \black CA does not account for \black inertial sensor \black biases, \black 
resulting in 
\black innacuracies \black in the \(\textbf{C}_b^n\) matrix's estimate. 
\black This is why \black the FA is implemented after the CA, 
\black i.e., aiming \black to improve \black 
the estimate of \(\textbf{C}_b^n\) by also estimating the sensors biases. In the \black traditional \black FA algorithm, a closed-loop \black EKF \black configuration is used. The first step consists of integrating a sub-set of the 
INS equations, \black such as\cite{titterton04}\black: 
$    \dot{\textbf{v}}^n=\textbf{C}_b^n \textbf{f}^b-(\textbf{w}_{en}^n+2\textbf{w}_{ie}^n)\times \textbf{v}^n+\textbf{g}^n$,
$    \dot{\textbf{C}}_b^n=\textbf{C}_b^n(\textbf{w}_{ib}^b\times)-(\textbf{w}_{in}^n\times)\textbf{C}_b^n$, \black
where \(\textbf{v}^n\) represents the Earth-related velocity vector in NED frame; the symbol × indicates the skew symmetric form of a vector; \(\textbf{w}_{en}^n\) is the transport rate vector, and 
\black $\textbf{w}_{in}^n=\textbf{w}_{ie}^n+\textbf{w}_{en}^n$.\black

The inputs to the \black 
INS \black equations are the \black inertial sensor \black measurement \black (denoted with $\tilde{}\;$) \black vectors 
\((\tilde{\textbf{f}}^b,\,\tilde{\textbf{w}}_{ib}^b)\). Since these measurements are corrupted by 
\black biases\black, the solution \black of the \black navigation equations exhibits error components that grow 
\black with \black time. The commonly adopted method to avoid this error 
\black accumulation \black is 
\black to use an EKF 
to estimate and compensate for the latter. 
By \black augmenting \black 
the inertial sensors biases as components \black of \black the state vector to be estimated, and \black sticking to \black the \black quasi-static \black 
condition, the prediction equation of the 
INS error propagation model is given by\cite{titterton04}:
\begin{equation} \label{eq:ins_error_model}
    \black\dot{\textbf{x}}
    =\begin{bmatrix}
        \black-2\textbf{w}_{ie}^n\times & \black-\textbf{g}^n\times & \textbf{I}_{3} & \bm{0}_{3} \\
        \textbf{A}_{\varphi v} &  \black-\textbf{w}_{ie}^n\times & \bm{0}_{3}  & -\textbf{I}_{3} \\
        \bm{0}_{3}  & \bm{0}_{3}  & \bm{0}_{3}  & \bm{0}_{3} \\
        \bm{0}_{3}  & \bm{0}_{3}  & \bm{0}_{3}  & \bm{0}_{3} \\
    \end{bmatrix}\black\textbf{x}
    +\begin{bmatrix}
        \textbf{C}_b^n & \bm{0}_{3} \\
        \bm{0}_{3}  & -\textbf{C}_b^n\\
        \bm{0}_{3}  & \bm{0}_{3} \\
        \bm{0}_{3}  & \bm{0}_{3} \\
    \end{bmatrix}\begin{bmatrix}
        \textbf{w}_a^b\\
        \textbf{w}_g^b\\
    \end{bmatrix},
\end{equation}
\black with $\textbf{x} = \left[\left(\delta\bm{v}^n\right)^T\;\bm{\varphi}^T\;\left(\textbf{b}_a^n\right)^T\;\left(\textbf{b}_g^n\right)^T\right]^T$ and: \black
\black
\begin{equation}
    \textbf{A}_{\varphi v}=\begin{bmatrix}
        0 & \frac{1}{R_{\lambda}+h} & 0 \\
        \frac{-1}{R_{L}+h} & 0 & 0\\
        0 & \frac{-\tan(L)}{R_{\lambda}+h} & 0\\
    \end{bmatrix},
\end{equation}
where \(\delta\bm{v}^n
    \black =\black\begin{bmatrix}
    \delta v_{N}&\delta v_{E}&\delta v_{D}
\end{bmatrix}^T\), \(\bm{\varphi}\black =\black\begin{bmatrix}
    \varphi_{N}&\varphi_{E}&\varphi_{D}
\end{bmatrix}^T\) are the \black $n$\black-frame-resolved 
INS velocity and misalignment errors; 
\(\textbf{b}_a^n
    \black =\black\begin{bmatrix}
    b_{aN}&b_{aE}&b_{aD}
\end{bmatrix}^T\), \(\textbf{b}_g^n
    \black =\black\begin{bmatrix}
    b_{gN}&b_{gE}&b_{gD}
\end{bmatrix}^T\) are the \black $n$\black -frame-resolved accelerometer and gyroscope biases; 
\(\textbf{w}_a^b
    \black =\black\begin{bmatrix}
    w_{ax}&w_{ay}&w_{az}
\end{bmatrix}^T\), \(\textbf{w}_g^b
    \black =\black\begin{bmatrix}
    w_{gx}&w_{gy}&w_{gz}
\end{bmatrix}^T\) are the \black $b$\black-frame-resolved accelerometer and gyroscope noises (supposed to be white and Gaussian), 
respectively; \black $\mathbf{I}_{3}$ and $\mathbf{0}_{3}$ \black are 3$\times$3 identity and zero matrices, also respectively; and \black \(R_{L}\) and \(R_{\lambda}\) are the \black meridian and transverse \black Earth radii of curvature. 

\black The process noise vector of \eqref{eq:ins_error_model} is characterized by the following covariance matrix: \black
$    \textbf{Q} = \text{diag}\left(\sigma_{wa}^2 \mathbf{I}_3,\, \sigma_{wg}^2 \mathbf{I}_3\right)$, \black
where \(\sigma_{wa}^2\) and \(\sigma_{wg}^2\) are the noise variances of the accelerometer and gyroscope signals, \black respectively\black.

The measurement equation, \black which \black relates the predicted state to the actual measurements observed in the system, 
\black is used to correct the state estimates 
during the EKF's update 
\black step. In the traditional approach, the measurement vector 
is defined 
\black as \black the difference between the \black 
INS-computed \black velocity vector~\((\hat{\textbf{v}}^n)\), 
and the true velocity, 
\black which, for \black a \black quasi-static \black 
condition, 
is assumed to \black have zero mean \black (
ZVU)\cite{silva14,silva18MEAS}: $\textbf{z}_{\text{\black ZVU}}=\hat{\textbf{v}}^n$, 
which is \black modeled \black as: \black
$    \textbf{z}_{\text{ZVU}}=\textbf{H}_{\text{ZVU}}\textbf{x} + \textbf{w}_{\text{ZVU}}$, \black
where \(\textbf{H}_{\text{\black ZVU}} = \begin{bmatrix} \black \textbf{I}_{3} & \black \bm{0}_{3}  & \black \bm{0}_{3}  & \black \bm{0}_{3}  \end{bmatrix}\) is the measurement matrix, 
\black and \(\textbf{w}_{\text{\black ZVU}}\) is the measurement noise vector, associated with \black the \black uncertainty in the 
\black ZVU \black assumption. This noise vector has the 
covariance matrix $\textbf{R}_{\text{\black ZVU}}=\sigma_{\text{ZVU}}^2\black \mathbf{I}_3 \black$, 
where \(\sigma_{\text{\black ZVU}}^2\) is the \black ZVU \black variance \black in all three axes. \black 

\section{The TRIAD-CBE Method} \label{sec:cbe}
TRIAD 
generates an estimated \black rotation \black matrix \(\hat{\textbf{C}}_b^n\), which is corrupted by errors. This matrix \black relates \black to the true matrix \(\textbf{C}_b^n\) by 
\black\cite{S2000}: $\hat{\textbf{C}}_b^n=(\textbf{I}+\textbf{E}_{ss}+\textbf{E}_s)\textbf{C}_b^n$, \black
with \black $\mathbf{E}_{ss} = -\bm{\varphi}\times$ and:
\begin{equation}
     \textbf{E}_s=
     \begin{bmatrix}
         \eta_N & o_D & o_E\\
        o_D & \eta_E & o_N\\
        o_E & o_N & \eta_D\\
     \end{bmatrix},
\end{equation}
where \(\textbf{E}_{ss}\) is a matrix associated with the \black mis\black alignment error vector (\(\bm{\varphi}\)), and \(\textbf{E}_s\) \black encompasses \black 
the normality \black and 
orthogonality \black error vectors \black(\(\bm{\eta} = [\eta_N\;\eta_E\;\eta_D]^T\) and $\mathbf{o}= [o_N\;o_E\;o_D]^T$). 

\black A first-order \black estimate of the normality/orthogonality error 
\black matrix \black 
can be obtained from the computed matrix \(\hat{\textbf{C}}_b^n\), \black namely \cite{silva25}:
$    \hat{\textbf{E}}_s=0.5\,[\hat{\textbf{C}}_b^n(\hat{\textbf{C}}_b^n)^T-I]$, whose elements \black
relate \black to some components of the inertial sensor \black biases\black\cite{silva18TIM}:
\black
\begin{equation} \label{eq:combined_expectations}
\begin{aligned}
    \mathbb{E}[\hat{\eta}_N]\black &= -\dfrac{\tan(L)}{g_D}b_{aN} + \dfrac{1}{w_e \cos(L)}b_{gN} + \tan(L)b_{L}, \\
    \mathbb{E}[\hat{\eta}_E]\black &= -\dfrac{\tan(L)}{g_D}b_{aN} - \dfrac{1}{g_D}b_{aD} + \dfrac{1}{w_e \cos(L)}b_{gN} \\
    &\quad + \tan(L)b_{L} - \dfrac{1}{g_D}b_{G}, \\
    \mathbb{E}[\hat{\eta}_D]\black &= -\dfrac{1}{g_D}b_{aD} -\dfrac{1}{g_D}b_{G}, \\
    \mathbb{E}[\hat{o}_E]\black &= -\dfrac{1}{2g_D}b_{aN} - \dfrac{\tan(L)}{2g_D}b_{aD} + \dfrac{1}{2w_e \cos(L)}b_{gD} \\
    &\quad + \dfrac{1}{2}b_{L} - \dfrac{\tan(L)}{2g_D}b_{G}.
\end{aligned}
\end{equation} \black
where \black $\mathbb{E}$ is the expectation operator, and \black
\(b_{L}\) and 
\black $b_{G}$ are the 
latitude and 
\black gravity biases, 
respectively\black. 

\black Assuming the CA is conducted at a place whose position and gravity acceleration are accurately known \black (which generally holds) \black allows us to employ the following bias estimation algorithms \black during the CA procedure (TRIAD-CBE)~\cite{silva18TIM}\black: 
\begin{equation} \label{eq:combined}
\begin{aligned}
    \hat{b}_{aD} &= -g_D \mathbb{E}[\hat{\eta}_D]\black, \\
    \hat{b}_{gN} &= w_e \cos(L) (\mathbb{E}[\hat{\eta}_E]-\mathbb{E}[\hat{\eta}_D]\black), \\
    \hat{b}_{gD} &= w_e \cos(L) (2 \mathbb{E}[\hat{o}_E]-\tan(L)\mathbb{E}[\hat{\eta}_D]\black).
\end{aligned}
\end{equation} \black

The 
\black claimed benefit of TRIAD-CBE 
\black is the possibility of pre-estimating the biases \black of 
\black \eqref{eq:combined} \black 
during the own \black CA (unfortunately, $b_{aN}$, $b_{aE}$, and $b_{gE}$ cannot be estimated \black likewise), \black which may 
accelerate \black (but not necessarily improve the accuracy of) \black the alignment process \black as a whole, i.e., CA+FA\black\cite{silva18TIM}.


\black
\section{The Optimized TRIAD-CBE Method} \label{sec:cbe-opt}

In 
TRIAD-CBE 
\black(reviewed \black in Section \ref{sec:cbe}), the biases are not estimated through an optimal procedure. In this Section, an optimized version of the method is presented \black(OPT-TRIAD-CBE)\black. To achieve this, 
\black \eqref{eq:combined_expectations} \black 
\black is \black 
firstly \black rewritten to include the noise components \black that corrupt the TRIAD-generated NON errors, which were ignored in 
\cite{silva18TIM}: \black 
\black
\begin{equation} \label{eq:combined_eta_o}
\begin{aligned}
    \eta_N &= \mathbb{E}[\hat{\eta}_N]-\dfrac{\tan(L)}{g_D}w_{aN} + \dfrac{1}{w_e \cos(L)}w_{gN} + \tan(L)w_{L}, \\ 
    \eta_E &= \mathbb{E}[\hat{\eta}_E]-\dfrac{\tan(L)}{g_D}w_{aN} - \dfrac{1}{g_D}w_{aD} \\
    &\quad + \dfrac{1}{w_e \cos(L)}w_{gN} + \tan(L)w_{L} - \dfrac{1}{g_D}w_{G}, \\ 
    \eta_D &= \mathbb{E}[\hat{\eta}_D] -\dfrac{1}{g_D}w_{aD} -\dfrac{1}{g_D}w_{G}, \\ 
    o_E &= \mathbb{E}[\hat{o}_E]-\dfrac{1}{2g_D}w_{aN} - \dfrac{\tan(L)}{2g_D}w_{aD} \\
    &\quad + \dfrac{1}{2w_e \cos(L)}w_{gD} + \dfrac{1}{2}w_{L} - \dfrac{\tan(L)}{2g_D}w_{G}. 
\end{aligned}
\end{equation} \black
where \(w_{a_N}, w_{a_D}, w_{g_N}, w_{g_D}\) are the \black north and down \black 
accelerometer and gyroscope noises, \black respectively, and $w_L$ and $w_G$ are the uncertainties in the latitude and gravity information, also respectively. \black 

Equations 
\black \eqref{eq:combined_expectations} 
\black and \black \eqref{eq:combined_eta_o} \black 
form a system of linear equations that relate the \black TRIAD-generated \black NON errors with the \black sensor \black 
biases \black and noises. \black This system, \black however, \black contains equations that are linearly dependent, 
\black namely, 
    $\eta_E=\eta_N+\eta_D$. \black
Therefore, the system is ill-conditioned, preventing the direct determination of all parameters (biases). \black In fact, only \black 
three linear combinations of the biases can be obtained, \black which are driven by the \black 
basis of the space of bias combinations that are linearly independent, \black i.e., 
estimable, namely: \black
\begin{equation}\label{eq:base}
    \text{base} = \begin{pmatrix}
            \begin{bmatrix}
                 0\\0\\1\\0\\0\\0\\1\\0
            \end{bmatrix},&
            \begin{bmatrix}
                \frac{-\tan(L)}{g_D}\\0\\0\\ \frac{1}{w_e \cos(L)}\\0\\0\\\tan(L)\\0
            \end{bmatrix},&
            \begin{bmatrix}
                \frac{-1}{2g_D}\\0\\0\\0\\0\\ \frac{1}{2 w_e \cos(L)}\\ \frac{1}{2}\\0
            \end{bmatrix}
        \end{pmatrix}.
    \end{equation}
    
\black From \black \eqref{eq:combined_expectations} \black 
and \eqref{eq:base}, one \black has that a \black possible selection of estimable linear combinations of the biases is: 
\begin{equation} \label{eq:b}
    \mathbf{b} = \left[\begin{array}{c}
         b_{aD}+b_G \\
         -\frac{\tan(L)}{g_D} b_{aN}+\frac{1}{w_e \cos(L)} b_{gN}+\tan(L) b_L \\
         -\frac{1}{2 g_D} b_{aN}+\frac{1}{2 w_e \cos(L)} b_{gD}+\frac{1}{2} b_L
    \end{array}\right].
\end{equation}

\black From 
\black \eqref{eq:combined_expectations}, 
\black \eqref{eq:combined_eta_o} \black 
and \black \eqref{eq:b}, 
it is possible to determine a system of equations that relates the \black NON errors to the \black estimable \black 
linear combinations of 
biases \black and system noises: \black 
$   \mathbf{z}_{\text{NON}} = \mathbf{H}_{\text{b}}
    \mathbf{b} + \mathbf{I}_{\text{NON}}\mathbf{w}_{\text{NON}}$, 
with \black $\mathbf{z}_{\text{NON}} = \left[\eta_N\;\eta_E\;\eta_D\;o_E\right]^T$, and:
\begin{equation}
    \black \mathbf{H}_{\text{b}}=\begin{bmatrix}
    0 & 1 & 0 \\
    \black -\frac{1}{g_D} & \black 1 & \black 0 \\
    \black -\frac{1}{g_D} & \black 0 & \black 0\\
    \black -\frac{\tan(L)}{2 g_D} & \black 1 & \black 1
    \end{bmatrix},
\end{equation}
\setlength{\arraycolsep}{0pt} 
\begin{equation*}
    \black \mathbf{I}_{\text{NON}}=\begin{bmatrix}
    -\frac{\tan(L)}{g_D} & 0 & \frac{1}{w_e\cos(L)} & 0 & \tan(L) & 0\\
    -\frac{\tan(L)}{g_D} & -\frac{1}{g_D} & \frac{1}{w_e\cos(L)} & 0 & \tan(L) & -\frac{1}{g_D}\\
    0 & -\frac{1}{g_D} & 0 & 0 & 0 & -\frac{1}{g_D}\\
    -\frac{1}{2g_D} & -\frac{\tan(L)}{2g_D} & 0 & \frac{1}{2w_e\cos(L)} & \frac{1}{2} & -\frac{\tan(L)}{2g_D}
    \end{bmatrix}
\end{equation*}
\setlength{\arraycolsep}{5pt} 
\begin{equation}
    \mathbf{w}_{\text{NON}} = \begin{bmatrix}
        w_{aN}&w_{aD}&w_{gN}&w_{gD}&w_{L}&w_G    
    \end{bmatrix}^T.
\end{equation}
  
 \black A solution to \black \eqref{eq:b}, 
 \(\hat{\mathbf{b}}\), can be obtained via a Weighted Least Squares (WLS) modified, via a whitening transformation, to account for singularities in the weighting matrix~\cite{golub2013matrix}, i.e., $\hat{\mathbf{b}} = ( \mathbf{\Sigma}^{-1} \mathbf{U}^T \mathbf{H}_{\text{b}} )^{\dagger} \, \mathbf{\Sigma}^{-1} \mathbf{U}^T \mathbf{z}_{\text{NON}}$, where $(\cdot)^{\dagger}$ denotes the Moore--Penrose pseudo-inverse, and $\mathbf{\Sigma}$ and $\mathbf{U}$ are obtained from the Singular Value Decomposition (SVD) of the mapping matrix $\mathbf{I}_{\text{NON}} \mathbf{W}^{1/2} = \mathbf{U} \mathbf{\Sigma} \mathbf{V}^T$, with $\mathbf{W} = \left(\mathbf{I}_{\text{NON}}\mathbf{R}_{\text{NON}}\mathbf{I}_{\text{NON}}^T\right)^{-1}$,  
 $   \mathbf{R}_{\text{NON}}= \text{diag}\left(\sigma_{w_{aN}}^2,\, \sigma_{w_{aD}}^2,\, \sigma_{w_{gN}}^2,\, \sigma_{w_{gD}}^2,\, \sigma_{w_{L}}^2,\, \sigma_{w_{G}}^2\right)$,     
 where $\sigma_i^2, i\in\{w_{aN},w_{aD},w_{gN},w_{gD},w_{L},w_{G}\}$, are the variances of the corresponding system noises.
 

\black Considering again the situation wherein the CA is conducted at a place whose position and gravity are accurately known (and following the same rationale observed in \cite{silva18TIM}), it is possible to recover optimized estimates for the down accelerometer bias, and north/down gyroscope biases by doing:
\begin{equation} \label{eq:b_hat2}
    \begin{bmatrix}
        \hat{b}_{aD} &
        \hat{b}_{gN} &
        \hat{b}_{gD}
    \end{bmatrix}^T = \begin{bmatrix}
        1 & w_e\cos(\hat{L}) & 2w_e\cos(\hat{L})
    \end{bmatrix}^T\mathbf{\hat{b}}.
\end{equation}

\black

With these modifications, 
TRIAD-CBE 
becomes an optimal estimator \black (OPT-TRIAD-CBE) \black for the inertial sensor biases. By compensating for these estimates \black into the raw sensor readings, it is straightforward to infer that the method should \black 
also provide 
optimal estimates 
\black for the misalignment \black errors. 

\black

\section{The Proposed Fine Alignment Method} \label{sec:prop}

\black As introduced in Section \ref{sec:intro}, in \black 
this paper, 
\black we \black also \black propose to augment the measurement vector of the traditional \black ZVU-based EKF \black FA 
with 
\black TRIAD-derived \black NON error estimates 
(\black TRIAD \black supposedly running in parallel with the FA), \black in order to see how the latter perform in the recognizably optimal 
KF-based estimator. 
Such an idea \black translates into the following augmented measurement equation for the EKF: 
\begin{equation}
    \textbf{z}=\begin{bmatrix}
        \textbf{z}_{\text{\black ZVU}} \\
        \textbf{z}_{\text{\black NON}} \\
    \end{bmatrix}\black\equiv\black\begin{bmatrix}
        \textbf{H}_{\text{\black ZVU}} \\
        \black \textbf{H}_{\text{NON}}' \\
    \end{bmatrix} \textbf{x}+\begin{bmatrix}
        \textbf{w}_{\text{\black ZVU}} \\
        \black \textbf{w}_{\text{NON}}' \\
    \end{bmatrix},
\end{equation}
with:
\setlength{\arraycolsep}{1pt}
\begin{equation*}
    \black \textbf{H}_{\text{NON}}'=\begin{bmatrix}
        \bm{0}_{1\times 6} & -\frac{\tan(L)}{g_D} & 0 & 0 & \frac{1}{w_e\cos(L)} & 0 & 0 \\
        \bm{0}_{1\times 6} & -\frac{\tan(L)}{g_D} & 0 & -\frac{1}{g_D} & \frac{1}{w_e\cos(L)} & 0 & 0 \\
        \bm{0}_{1\times 6} & 0 & 0 & -\frac{1}{g_D} & 0 & 0 & 0 \\
        \bm{0}_{1\times 6} & -\frac{1}{2g_D} & 0 & -\frac{\tan(L)}{2g_D} & 0 & 0 & \frac{1}{2w_e\cos(L)} \\
    \end{bmatrix}.
\end{equation*}
\setlength{\arraycolsep}{5pt}

\black The covariance matrix of the \black augmented \black noise vector is given by $\textbf{R}= \text{\black diag}\left(\textbf{R}_{\text{\black ZVU}},\, \black\textbf{R}_{\text{NON}}'\black\right)$ \black with: \black
\begin{equation} \label{eq:R}
\begin{aligned}
    \textbf{R}_{\text{NON}}' &= \text{ diag}\left(\sigma_{w_{aN}}^2,\,\sigma_{w_{aD}}^2,\,\sigma_{w_{gN}}^2,\,\sigma_{w_{gD}}^2\right) \\ 
    &\approx \text{diag}\left(\sigma_{wa}^2,\,\sigma_{wa}^2,\,\sigma_{wg}^2,\,\sigma_{wg}^2\right),
\end{aligned}
\end{equation}
\black where $\sigma_{wa}^2$ and $\sigma_{wg}^2$ are the same variances defined 
\black in Section \ref{sec:trad}\black. \black Hereinafter, this augmented EKF-based FA model will be referred to as ZVU+NON.\black


\section{Simulated Experiments} \label{sec:res}

To evaluate the performance of the 
\black proposed \black methods (OPT-TRIAD-CBE and ZVU+NON), 
\black simulated tests were carried out. First, 
\black aiming at \black a quantitative analysis of the performance of 
\black the latter \black w.r.t. \black the traditional 
\black approaches (TRIAD-CBE and \black ZVU)\black, a single-run simulation was executed. 
For this test, a stationary inertial sensor data array was generated with one hour of duration, sampled at a frequency of 100 Hz, and considering 
body and navigation 
frames \black perfectly \black aligned. 
The inertial sensor measurements were corrupted by \(0.5\) mg and \(0.5\) deg/h of constant biases, \black as well as \black 
by white noises with 
\black S\black tandard \black D\black eviations (SDs) \black of \(0.1\) mg and \(0.1\) deg/h. \black For 
\black benchmarking \black purposes, ZVU+CERGU-based FA \cite{silva14,silva18MEAS} 
\black was \black also deployed throughout the simulation. During the first minute of the \black data set, \black 
ordinary TRIAD was deployed to initialize \black all FA approaches. \black 

\begin{figure}[t!] 
    \centering
    \includegraphics[width=\columnwidth]{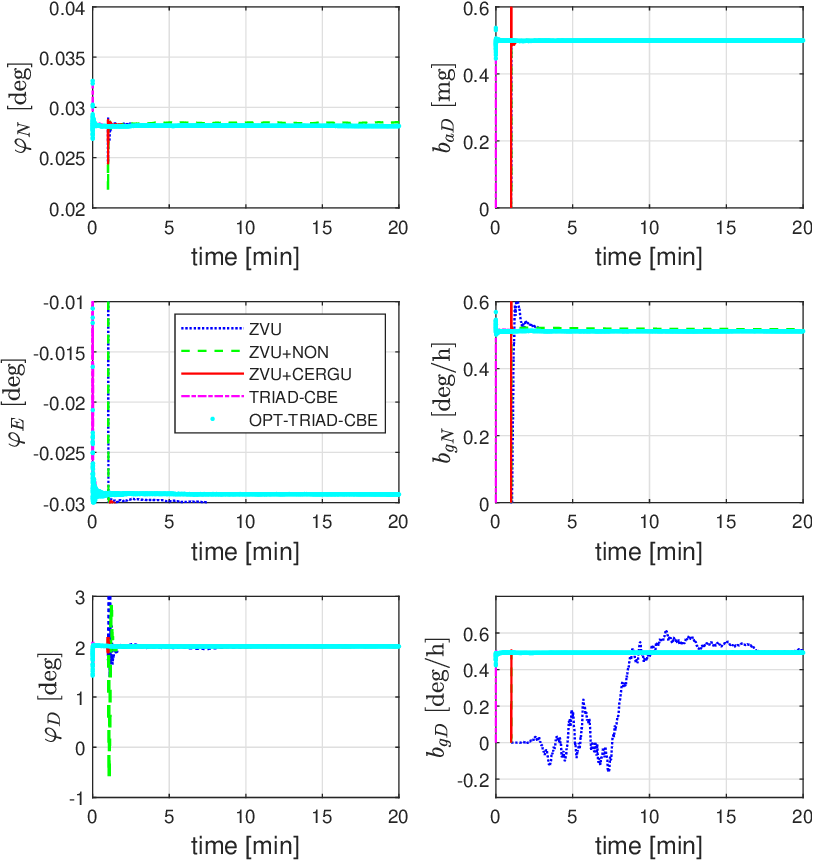} 
    \caption{\black Misalignment and bias estimates
    (single-run simulated experiment).}
    \label{fig:align_bias_sim}
\end{figure}

\black Figure \ref{fig:align_bias_sim} presents the estimates of the misalignment errors and the \black \(b_{aD}, b_{gN},\) and \(b_{gD}\) biases components obtained by the 
methods \black under evaluation.
As can be seen, there are no significant differences in terms of misalignment accuracies. 
From the bias estimation standpoint, none of the investigated FA methods solved \black the 
\black \black estimability~\cite{baram88,goshen-meskin92} \black issue of the horizontal accelerometer and east gyroscope 
\black biases (reason why their estimates have not been plotted), which are known to constrain the accuracy of the \black quasi-static \black 
self-alignment problem \cite{SILVA201745,silva17}). Nevertheless, 
ZVU+NON-based FA, ZVU+CERGU-based FA, and specially TRIAD-CBE \black and OPT-TRIAD-CBE \black largely improved \black the 
convergence rate of the 
\black remaining 
\black ``estimable" \black bias \black components 
(from 30 minutes to few seconds \black for instance, for $b_{gD}$)\black.
\begin{table}[b!]
\centering
\caption{Stochastic Variables 
(MC \black Simulation)}
\label{tab:mc_par}
\begin{tabular}{c|c|c}  
\textbf{Variables} & \textbf{PDF} & \textbf{PDF parameters} \\
\hline
\black Accelerometer \black biases [mg] & Normal & \(\mu\)= 0, \(\sigma\)= 0.5 \\
\black Gyroscope \black biases [deg/h] & Normal  & \(\mu\)= 0, \(\sigma\)= 0.5 \\
\end{tabular}
\end{table}

\begin{figure}[t!]
    \centering
    \includegraphics[width=\columnwidth]{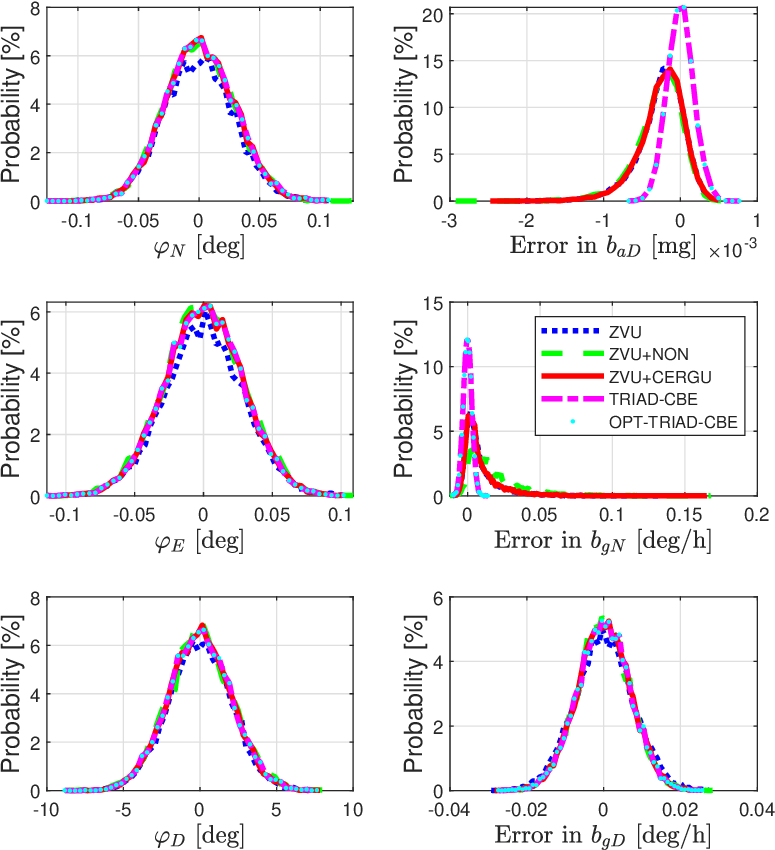} 
    \caption{Misalignment and biases error distributions 
    (MC \black simulation).}
    \label{fig:align_bias_mc}
\end{figure}

From the results of \black a \black single-run simulation, \black conclusion about the 
misalignment/bias estimation performance (in terms of accuracy/convergence rate) \black of the proposed methods \black may be misleading\black. 
For a \black more reliable assessment of \black the 
\black latter\black, a 
\black MC \black simulation was \black deployed\black. For this analysis, 10,000 statistically independent realizations of the \black previous \black simulation were repeated. The stochastic variables adopted in this simulation and their statistical models are summarized in Table \ref{tab:mc_par}, \black where PDF stands for \black P\black robability
\black Density \black F\black unction\black. In each run, the last values obtained for the misalignment and \black ``estimable" bias \black errors were saved and used as data for the \black MC \black analysis. The first column of Fig. \ref{fig:align_bias_mc} presents the 
\black PDF estimates \black for the components of the misalignment vector \black obtained \black  from the MC simulation\black. 
\black As can be seen, all of the investigated \black methods show \black negligible \black 
differences (smaller than \(10^{-2}\) deg) in terms of accuracy/precision. This can be better observed in Table 
\ref{tab:mc_mean_sd}, \black which summarizes \black 
the mean and SDs of the estimated PDFs, \black rounded according to Chernoff's criteria \cite{chernoff1952measure}\black. 

\black Aiming \black to validate \black 
these results, \black analytical \black expressions for the 
\black SDs \black (hereinafter denoted by $\sigma$) \black were obtained for each of the misalignment error components. These equations were derived from
\black\cite{silva16} for TRIAD-CBE \black and OPT-TRIAD-CBE\black, and from \cite{SILVA201745,silva17} for the FA methods\black. 
\black Consistently, the SD \black expressions are the same for all 
\black of the investigated \black methods, and are given by:
\black 
\begin{equation} \label{eq:combined_sigma_phi}
\begin{aligned}
    \sigma_{\varphi_n} &= \frac{\sigma_{b_{aE}}}{g_D} = 0.029 \;\text{deg}, \\
    \sigma_{\varphi_E} &= \frac{\sigma_{b_{aN}}}{g_D} = 0.029 \;\text{deg}, \\
    \sigma_{\varphi_D} &= \sqrt{\left(\frac{\tan(L)\sigma_{b_{aE}}}{g_D}\right)^2+\left(\frac{\sigma_{b_{gE}}}{w_e \cos(L)}\right)^2} \\
    &= 2.067 \;\text{deg}.
\end{aligned}
\end{equation} \black
\begin{table}[t!]
\centering
\caption{\black Mean and SDs of the misalignment errors 
(MC simulation)}
\label{tab:mc_mean_sd}
\begin{tabular}{c|c|c|c}  
\textbf{Method / Error} & \(\varphi_N\) [deg] & \(\varphi_E\) [deg] & \(\varphi_D\) [deg] \\
\hline
\multirow{2}{*}{\textbf{ZVU}} & {0.000} & {0.000} & -0.010  \\
             & $\pm$ {0.029} & $\pm$ {0.029} & $\pm$ {2.036} \black \\
\multirow{2}{*}{\textbf{ZVU+NON}} & {0.000} & {0.000} & -0.004  \\
                 & $\pm$ {0.029} & $\pm$ {0.029} & $\pm$ 2.151 \\
\multirow{2}{*}{\textbf{ZVU+CERGU}} & {0.000} & {0.000} & -0.0013 \\
                   & $\pm$ {0.029} & $\pm$ {0.029} & $\pm$ 2.060 \\
\multirow{2}{*}{\textbf{TRIAD-CBE}} & {0.000} & {0.000} & -0.0015  \\
                   & $\pm$ {0.029} & $\pm$ {0.029} & $\pm$ 2.075 \\
\black \multirow{2}{*}{\textbf{OPT-TRIAD-CBE}} & {0.000} & {0.000} & -0.0015 \\
                   & \black $\pm$ {0.027} & \black $\pm$ {0.027} & \black $\pm$ 2.070 \\
\end{tabular}
\end{table}
The values obtained from the analytical expressions \black of \eqref{eq:combined_sigma_phi} \black 
agree \black with the \black distributions from Table \ref{tab:mc_mean_sd}\black. 
This confirms that, 
\black from \black the misalignment error components \black estimation standpoint, 
all of the investigated methods have \black equivalent performance, \black i.e., the FA methods were not able to improve over the CA accuracy, differently from what is sometimes claimed in the literature\black.
\begin{table}[b!]
\centering
\caption{\black Mean and SDs of the bias errors 
(MC simulation)}
\label{tab:mc_bias}
\begin{tabular}{c|c|c|c}  
\textbf{Method / Error} & \(b_{aD}\) [mg] & \(b_{gN}\) [deg/h] & \(b_{gD}\) [deg/h] \\
\hline
\multirow{2}{*}{\textbf{ZVU}} & -0.0002  & \black 0.009  & \black 0.000 \\
 & $\pm$ 0.0003 & $\pm$\black 0.013 & $\pm$\black 0.008 \\
\multirow{2}{*}{\textbf{ZVU+NON}} & -0.0002  & \black 0.008 & \black {0.000} \\
 & $\pm$ 0.0003 & $\pm$\black 0.019 & $\pm$\black {0.007} \\
\multirow{2}{*}{\textbf{ZVU+CERGU}} & 0.0003 & \black 0.009 & \black {0.000} \\
 & $\pm$ 0.0003 & $\pm$\black 0.013 & $\pm$\black {0.007} \\
\multirow{2}{*}{\textbf{TRIAD-CBE}} & {0.0000} & \black {0.000} & \black {0.000} \\
 & $\pm$ {0.0001} & $\pm$\black {0.003} & $\pm$\black {0.007} \\
\black \multirow{2}{*}{\textbf{OPT-TRIAD-CBE}} & \black {0.0000} & \black {0.000} & \black {0.000} \\
 & \black $\pm$ {0.0001} & \black $\pm$\black {0.003} & \black $\pm$\black {0.007} \\
\end{tabular}
\end{table}

The estimated PDF for the 
bias \black error \black components are also presented in Fig. 
\ref{fig:align_bias_mc}. 
As can be observed, 
the methods 
\black under evaluation showed \black similar performances (in terms of accuracy and precision) only \black for \black the estimation of the 
\black down component \black of the gyro bias. 
\black With respect to the down accelerometer bias and north gyro bias, TRIAD-CBE \black and OPT-TRIAD-CBE were \black 
the methods that performed best \black(without noticeable differences between them)\black, which can be \black better observed 
\black from \black the \black metrics \black 
presented in 
\black Table \ref{tab:mc_bias}, \black also rounded according to Chernoff's criteria~\cite{chernoff1952measure}. \black 

As \black done \black 
\black for \black the misalignment errors, to validate the results presented in Table \ref{tab:mc_bias}, \black analytical \black expressions for the 
\black SDs \black of the bias components were obtained. For TRIAD-CBE, 
the expressions, \black which are based on 
\black \eqref{eq:combined}, \black 
were obtained from \cite{silva16}, 
and are given by:
\black
\begin{equation} \label{eq:combined_sigma_b}
\begin{aligned}
    \sigma_{b_{aD}} &= g_D\sigma_{\eta_D}, \\
    \sigma_{b_{gN}} &= \sqrt{\left[w_e \cos(L) \sigma_{(\eta_E-\eta_D)}\right]^2+\left[\frac{w_e \sin(L) \sigma_{b_{aN}}}{g_D}\right]^2}, \\
    \sigma_{b_{gD}} &= \bigg\{\left[w_e \cos(L) \sigma_{(2o_E-\tan(L)\eta_D)}\right]^2 \\
    &\quad +\left[\frac{w_e \cos(L) \sigma_{b_{aN}}}{g_D}\right]^2\bigg\}^{0.5}.
\end{aligned}
\end{equation} \black

\black The 
\black SDs \black for each of the 
NON error \black linear combinations of \black \eqref{eq:combined_sigma_b} \black 
were \black obtained experimentally; the last value of each 
\black MC realization was \black stored, and at the end of the simulation, the 
\black SD was \black calculated. 
\black Following this approach, \black the values obtained for the bias uncertainties were: \black $\sigma_{b_{aD}}= 0.0001$ mg, $\sigma_{b_{gN}}= 0.003$ deg/h, and $\sigma_{b_{gD}}= 0.007$ deg/h, which \black are in agreement with the \black metrics \black 
\black from Table \ref{tab:mc_bias}, \black confirming the precision of 
\black TRIAD-CBE \black in estimating the bias components. \black With respect to OPT-TRIAD-CBE, analytical expressions for the bias dispersions were obtained via direct application of the standard linear transformation rule for covariances over~\eqref{eq:b_hat2}, with $\text{cov}(\hat{\mathbf{b}}) = \left(\mathbf{H}_{\text{b}}^T\mathbf{W}\mathbf{H}_{\text{b}}\right)^{-1}$. In doing so, the outcomes (theoretical SDs) proved to be 
the same as those from TRIAD-CBE, which also corroborates the results from Table \ref{tab:mc_bias}. \black
%

For the evaluation of the 
\black precision of the FA methods in estimating the biases, corresponding \black expressions for the 
\black their SDs \black were obtained from 
\cite{SILVA201745}, \black namely:
\black
\small
\black
\begin{equation} \label{eq:sigma_baD2}
\begin{aligned}
    \sigma_{b_{aD}} &= \sqrt{(\sigma_{\delta\dot{v}_D})^2 + [2w_{e}\cos(h)\sigma_{\delta v_E}]^2},
\end{aligned}
\end{equation}
\begin{multline} \label{eq:sigma_bgN}
    \sigma_{b_{gN}} = \bigg\{\left(\dfrac{\sigma_{\delta\ddot{v}_E}}{g_D}\right)^2 + \left(\dfrac{3w_e\sin(L)\sigma_{\delta\dot{v}_N}}{g_D}\right)^2 \\
    + \left(\dfrac{2w_e\cos(L)\sigma_{\delta\dot{v}_D}}{g_D}\right)^2 + \left[\left(\dfrac{1}{R_{\lambda} + h} - \dfrac{2w_e\sin^2(L)}{g_D}\right)\sigma_{\delta{v}_E}\right]^2 \\
    + \left(\dfrac{w_e\sin(L)\sigma_{b_{aN}}}{g_D}\right)^2\bigg\}^{0.5},
\end{multline}
\begin{multline} \label{eq:sigma_bgD}
    \sigma_{b_{gD}} = \bigg\{\left(\dfrac{\sigma_{\delta\dddot{v}_N}}{g_Dw_e\cos(L)}\right)^2 + \left(\dfrac{3\tan(L)\sigma_{\delta\ddot{v}_E}}{g_D}\right)^2 \\ 
    + \left[\left(\dfrac{2w_e\sin^2(L)}{g_D\cos(L)} - \dfrac{1}{w_e\cos(L)(R_L + h)} - \dfrac{w_e\cos(L)}{g_D}\right)\sigma_{\delta\dot{v}_N}\right]^2 \\ 
    + \left(\dfrac{2w_e\sin(L)\sigma_{\delta\dot{v}_D}}{g_D}\right)^2 + \left[\left(\dfrac{\tan(L)}{R_{\lambda} + h} + \dfrac{w_e^2\sin(2L)}{g_D}\right)\sigma_{\delta{v}_E}\right]^2 \\
    + \left(\dfrac{w_e\cos(L)\sigma_{b_{aN}}}{g_D}\right)^2\bigg\}^{0.5}.
\end{multline}
\normalsize

\black As can be seen from \eqref{eq:sigma_baD2}, for \black the particular case of \(b_{aD}\) \black (estimated from the FA methods), its precision is only \black function of 
\black the velocity errors and their derivatives, which are states with ``high \black estimability"\black\cite{baram88,goshen-meskin92,silva17}, \black as they are directly \black measured \black 
via the ZVUs\black. Therefore, it is expected that \(\sigma_{b_{aD}}\) will 
\black be a small \black value, which is confirmed 
\black from \black the \black metrics \black 
in Table \ref{tab:mc_bias}. For the other two bias components of 
\black \eqref{eq:sigma_bgN}-\eqref{eq:sigma_bgD}, \black 
\black if the uncertainties in the velocity errors and their derivatives \black are \black similarly \black disregarded, this leads us to\black: 
\begin{equation} \label{eq:fa_bgN}
\begin{aligned}
    \sigma_{b_{gN}} &\approx \frac{w_e \sin(L)}{g_D}\sigma_{b_{aN}} = 0.003\;\text{deg/h}, \\
    \sigma_{b_{gD}} &\approx \frac{w_e \cos(L)}{g_D}\sigma_{b_{aN}} = 0.007\;\text{deg/h}.
\end{aligned}
\end{equation}

As can be observed, the \black analytically \black calculated value for \(\sigma_{b_{gD}}\) \black agrees \black with the one obtained from the simulation (Table \ref{tab:mc_bias}). However, this is not the case for \(\sigma_{b_{gN}}\). This discrepancy may have been caused by the assumption made in determining the \black analytical \black expression for the \black SD\black, \black i.e., neglecting the uncertainty of \black 
states with ``high \black estimability". 
\black As can be seen in \black \eqref{eq:sigma_bgN}-\eqref{eq:sigma_bgD}, \black 
the uncertainties in the FA estimates of the north and down gyro biases are functions of \black 
higher-order derivatives of the velocity errors. \black Even though the velocity errors are \black estimable \black 
states in the FA alignment problem, there will always be a Cramer Rao \black lower \black bound 
defining their minimal uncertainties. It is also well known that the differentiation of noisy variables/states with respect to short time intervals amplifies the noise \cite{brown12}. Such amplified noise/uncertainty of the velocity error derivatives hence, once multiplied by the corresponding coefficients of \black \eqref{eq:sigma_bgN}, \black 
might justify why \eqref{eq:fa_bgN} did not agree with the SDs empirically determined for the north gyro bias (Table \ref{tab:mc_bias}). 

To confirm this, \black the same strategy previously adopted for TRIAD-CBE was \black 
employed, \black i.e., \black at the end of each \black MC \black realization, the \black velocity errors were stored, their derivatives computed, and by the completion of the simulation, the corresponding SDs were \black experimentally \black calculated and used into \black \eqref{eq:sigma_bgN}. \black 
For all of the three FA methods under investigation, i.e., ZVU-, ZVU+NON-, and ZVU+CERGU-based EKFs, the ``analytically" obtained SD for the north gyro bias was 
    $\sigma_{b_{gN}}= 0.011 \;\text{deg/h}$,
which confirms the outcomes \black 
from the MC simulation (Table \ref{tab:mc_bias}).
\begin{table}[t!]
\centering
\caption{
Convergence time 
\black mean values, in seconds (MC simulation)}
\label{tab:time}
\begin{tabular}{c|c|c|c|c|c|c}  
\textbf{Method / Error} & \(\varphi_N\) & \(\varphi_E\) & \(\varphi_D\) & \(b_{aD}\) & \(b_{gN}\) & \(b_{gD}\) \\
\hline
\textbf{ZVU} & 339 & 296 & 338 & 144 & 183 & 1742 \\
\textbf{ZVU+NON} & 263 & 247 & 18 & 69 & 76 & 77 \\
\textbf{ZVU+CERGU} & 294 & 312 & 83 & 86 & 98 & 96 \\
\textbf{TRIAD-CBE} & 34 & 25 & 27 & 24 & 27 & 36 \\
\black \textbf{OPT-TRIAD-CBE} & \black 37 & \black 38 & \black 55 & \black 36 & \black 35 & \black 45 \\
\end{tabular}
\end{table}

Another parameter considered in the evaluation of the \black investigated \black methods 
\black was \black the convergence 
\black rate\black, defined as the time \black the estimate takes to \black 
stabilize within a range of \(\pm5\%\) of 
\black its steady state \black value. 
\black Table \ref{tab:time} summarizes the MC-derived \black mean \black and SDs
\black of the \black misalignment/bias convergence rates 
obtained from the methods under evaluation
\black. As can be seen, 
\black ZVU+NON-based FA significantly \black reduced the convergence time of the estimates (from minutes to a few seconds), \black even w.r.t. ZVU+CERGU-based FA, which was defined as a benchmark method. Particularly \black impressive results were obtained \black by TRIAD-CBE \black and OPT-TRIAD-CBE, \black which \black were \black 
able to stabilize all states of interest in no more than \black 29 and 41 \black 
seconds (in average), respectively. \black  


\section{Real-World Experiments} \label{sec:exp}

\begin{figure}[t!]
    \centering
    \includegraphics[width=\columnwidth]{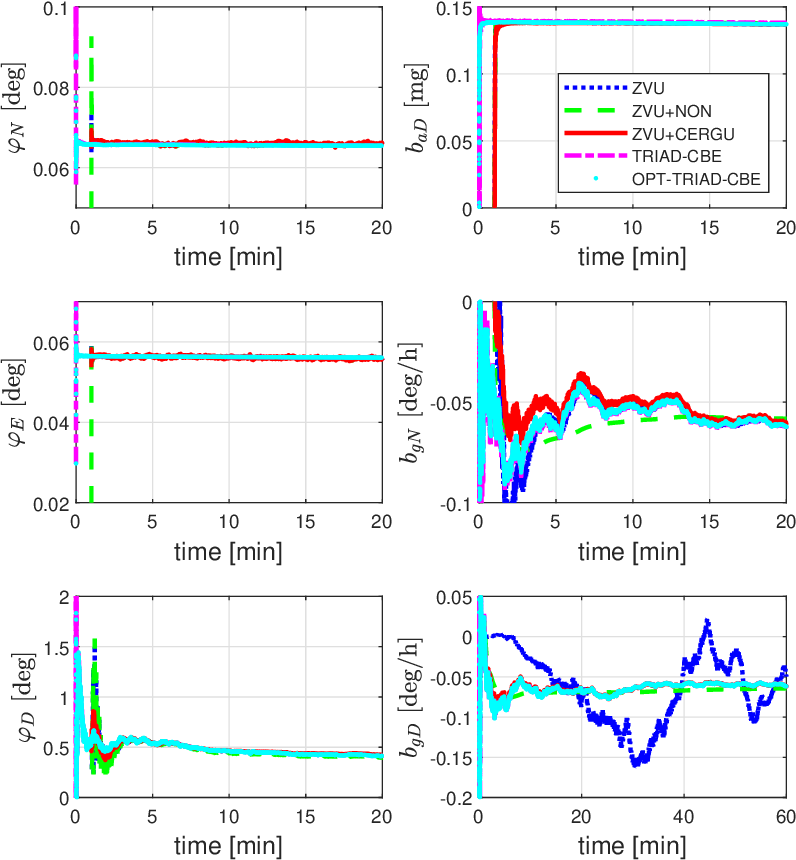} 
    \caption{Misalignment and bias estimates 
    \black (turntable \black 
    experiment).}
    \label{fig:misalign}
\end{figure}

\begin{table}[b!] \black
\centering
\caption{\black IMU technical specifications}
\label{tab:Primus300_spec}
\setlength{\tabcolsep}{3pt} 
\begin{tabular}{c|c|c|c|c}  
\multirow{2}{*}{\textbf{Parameter}} & \multicolumn{2}{c|}{\textbf{TNL-16G}} & \multicolumn{2}{c}{\textbf{Primus}} \\
\cline{2-5}
& \textbf{Gyro.} & \textbf{Accel.} & \textbf{Gyro.} & \textbf{Accel.} \\
\hline
Range [deg/s or g] & 600 & 40 & 150 & 2 \\
Bias [deg/h or mg] & 0.1 & 0.1 & 0.02 & 1 \\
Scale factor [ppm] & 70 & 100 & 20 & 600 \\
Noise density [deg/\(\sqrt{\text{h}}\)] & 0.002 & (not specified) & 0.002 & (not specified) \\
\end{tabular}
\end{table}
\black To reproduce the results obtained in the simulated test, two \black real-world experiments \black 
were conducted, one using a commercial intermediate-grade \black TNL-16G Inertial Measurement Unit (IMU) from Israel Aerospace Industries, and a second using a \black true north finder 
Primus from Safran Electronics \& Defense 
(Table \ref{tab:Primus300_spec}). \black As for the first test setup, \black the IMU was mounted aligned to the navigation frame on a \black high-end \black three-axis turntable. 
The misalignment errors and bias 
estimates obtained by the methods under evaluation are presented \black in Fig. \ref{fig:misalign}. \black As can be observed, the 
\black latter \black exhibited the same patterns observed in Section \ref{sec:res} 
(Fig. \ref{fig:align_bias_sim}), \black namely: (a) The accuracy/precision of \black the misalignments \black is the same regardless of the method (CA or FA); and (b) the proposed \black methods (specially OPT-TRIAD-CBE) \black 
significantly speed up the convergence rate of both misalignments and inertial sensor biases, compared to ZVU- and ZVU-CERGU-based FAs. \black 
This can be \black further \black confirmed from the \black metrics \black 
in Tables \ref{tab:re_mean_sd}-\ref{tab:re_bias}.

\begin{figure}[t!]
    \centering
    \includegraphics[width=0.4\textwidth]{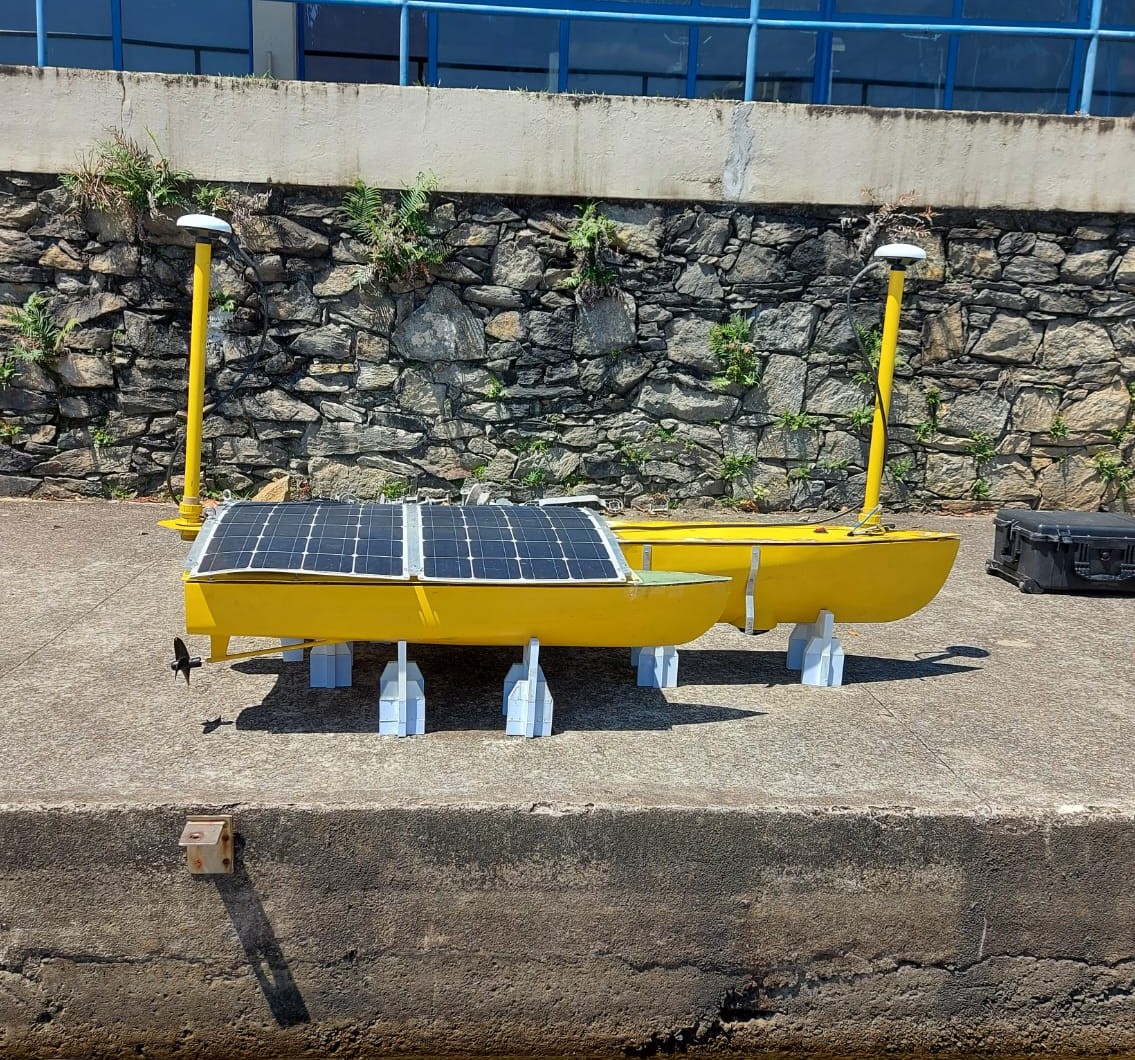} 
    \caption{\black ASV \black 
    on the pier.}
    \label{fig:asv_pier}
\end{figure}

\begin{figure}[t!]
    \centering
    \includegraphics[width=\columnwidth]{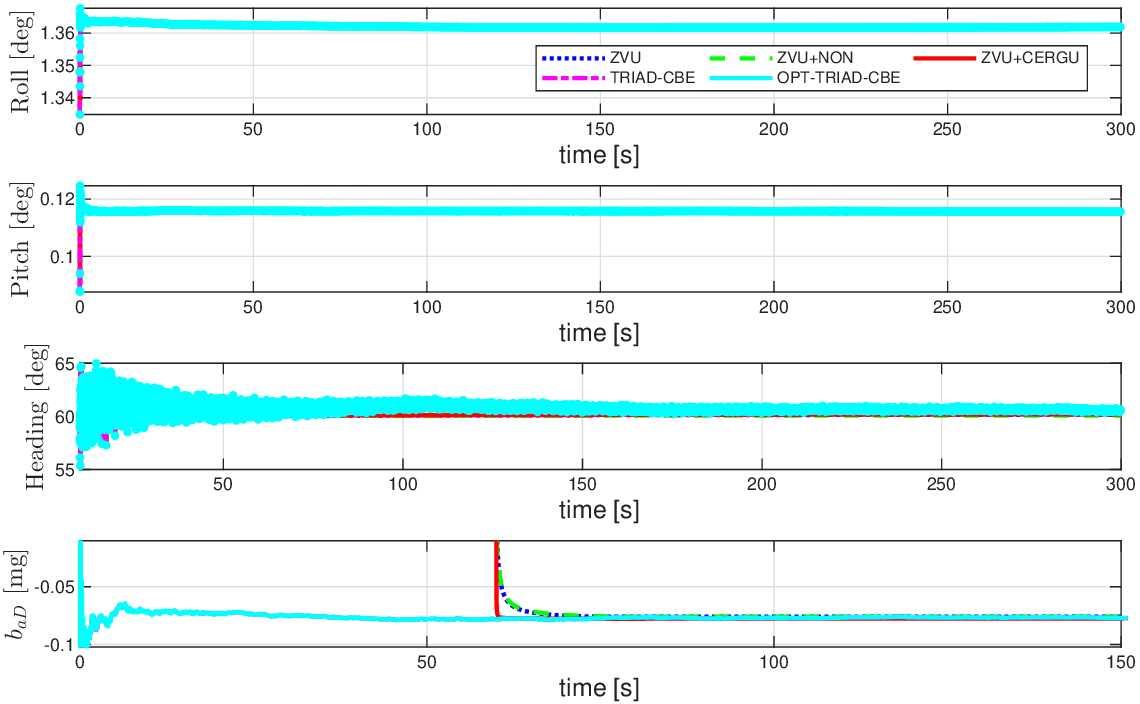} 
    \caption{Attitude and down accelerometer bias estimates 
    (ASV \black 
    experiment).}
    \label{fig:attit2}
\end{figure}

\begin{figure}[t!]
    \centering
    \includegraphics[width=\columnwidth]{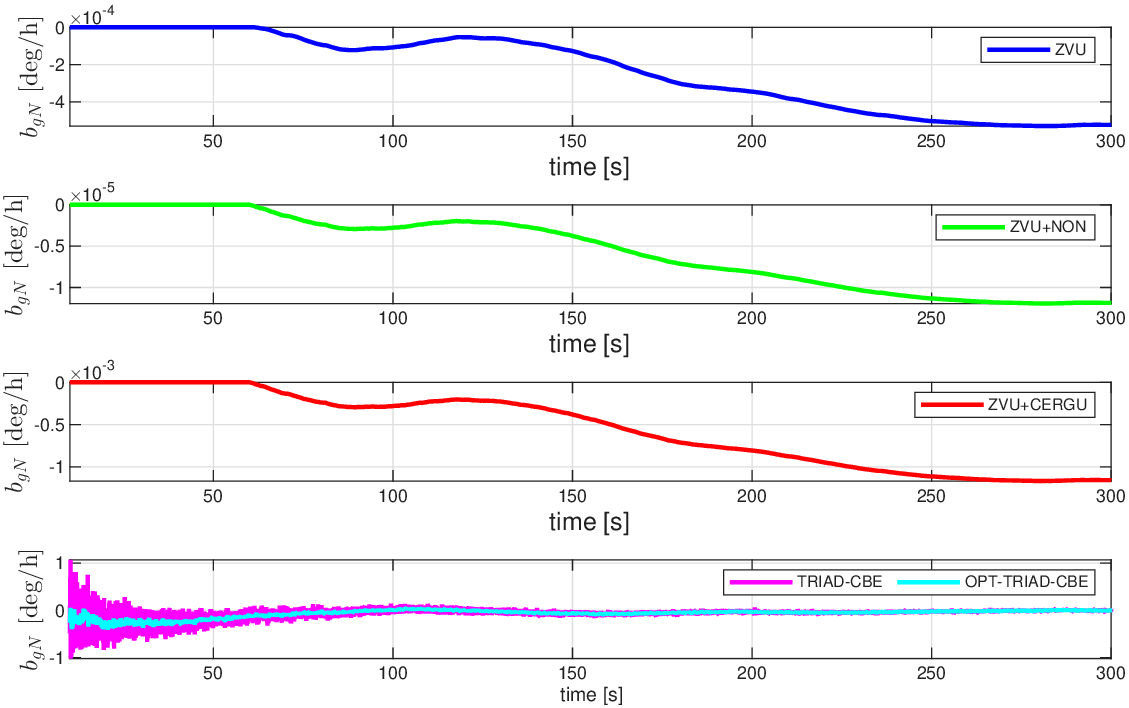} 
    \caption{
    \black North gyroscope bias \black estimates 
    (ASV \black 
    experiment).}
    \label{fig:estbgN2}
\end{figure}

\begin{figure}[t!]
    \centering
    \includegraphics[width=\columnwidth]{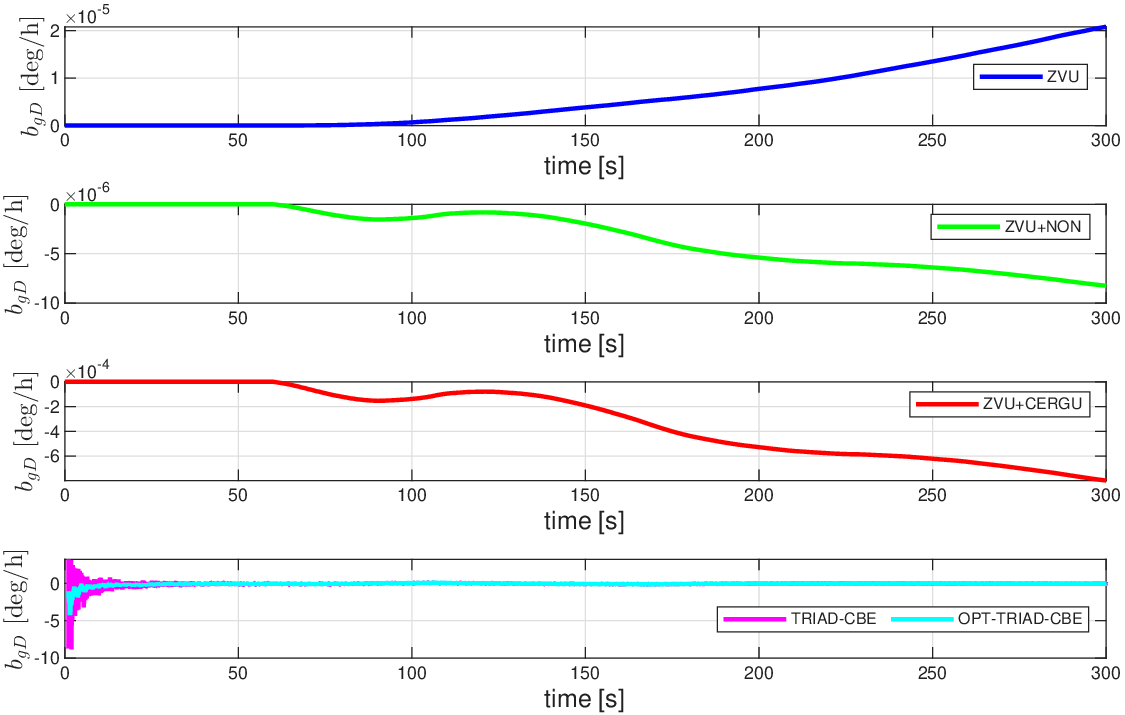} 
    \caption{
    \black Down gyroscope bias \black estimates 
    (ASV \black 
    experiment).}
    \label{fig:estbgD2}
\end{figure}

\black The second experimental test was carried out using an 
ASV, 
illustrated in Fig. \ref{fig:asv_pier}, which was designed as a test platform for the development of \black AUV \black navigation algorithms. 
The test was carried out 
\black at \black the olympic rowing course of the University of São Paulo, 
with the \black ASV sitting still 
on a concrete pier.
\black The attitude and biases estimated by the 
methods under analysis are presented in Figs. \ref{fig:attit2}-\ref{fig:estbgD2}. 
\begin{table}[b!]
\centering
\caption{Mean and SDs of the misalignment errors 
(turntable \black 
experiment)}
\label{tab:re_mean_sd}
\begin{tabular}{c|c|c|c}  
\textbf{Method / Error} & \(\varphi_N\) [deg] & \(\varphi_E\) [deg] & \(\varphi_D\) [deg] \\
\hline
\textbf{ZVU} & \(0.066\pm8\text{E-}5\)  & \(0.055\pm7\text{E-}5\)  & \(0.425\pm1\text{E-}3\) \\
\textbf{ZVU+NON} & \(0.066\pm8\text{E-}5\)  & \(0.055\pm7\text{E-}5\)  & {\(0.392\pm4\text{E-}4\)} \\
\textbf{ZVU+CERGU} & \(0.066\pm9\text{E-}5\)  & \(0.055\pm9\text{E-}5\)  & \(0.420\pm1\text{E-}4\) \\
\textbf{TRIAD-CBE} & {\(0.065\pm3\text{E-}6\)}  & {\(0.055\pm4\text{E-}6\)}  & \(0.403\pm1\text{E-}3\) \\
\black \textbf{OPT-TRIAD-CBE} & \black{\(0.066\pm3\text{E-}6\)}  & \black{\(0.055\pm5\text{E-}6\)}  & \black\(0.403\pm1\text{E-}3\) \\
\end{tabular}
\end{table}
\begin{table}[b!]
\centering
\caption{Mean and SDs of the bias errors 
(turntable \black 
experiment)}
\label{tab:re_bias}
\begin{tabular}{c|c|c|c}  
\textbf{Method / Error} & \(b_{aD}\) [mg] & \(b_{gN}\) [deg/h] & \(b_{gD}\) [deg/h] \\
\hline
\textbf{ZVU} & \(0.13\pm3\text{E-}4\)  & \(-0.05\pm3\text{E-}4\)  & \(-0.07\pm2\text{E-}2\) \\
\textbf{ZVU+NON} & \(0.13\pm1\text{E-}4\)  & \(-0.05\pm1\text{E-}4\)  & \(-0.06\pm2\text{E-}4\) \\
\textbf{ZVU+CERGU} & \(0.13\pm3\text{E-}4\)  & \(-0.05\pm3\text{E-}4\)  & \(-0.06\pm1\text{E-}3\) \\
\textbf{TRIAD-CBE} & \(0.13\pm3\text{E-}4\)  & \(-0.05\pm5\text{E-}4\)  & \(-0.06\pm1\text{E-}3\) \\
\black \textbf{OPT-TRIAD-CBE} & \black\(0.13\pm4\text{E-}4\)  & \black\(-0.05\pm6\text{E-}4\)  & \black\(-0.06\pm1\text{E-}3\) \\
\end{tabular}
\end{table} 

\black As can be seen, \black the estimates of the 
methods converged to values with differences less than one hundredth of a degree for roll and pitch, and \black less than one \black 
tenth for heading. This \black very same \black behavior was observed in the results of the simulations and the previous experiment (Figs. \ref{fig:align_bias_sim} and \ref{fig:misalign}). \black With respect to \black 
the estimation of the accelerometer bias (\(b_{aD}\)), despite the shorter \black alignment \black 
time (300 s) compared to the previous simulations and experiment (3600 s), all 
methods were able to estimate \black the latter \black 
with differences smaller than \(10^{-4}\) m/s², due to the high \black estimability \black 
of this bias component \cite{baram88,goshen-meskin92,silva17}. \black Likewise for the \black 
attitude estimates, this behavior \black has been \black 
previously observed (Figs. \ref{fig:align_bias_sim} and \ref{fig:misalign}). 

\black Concerning \black 
the estimation of the gyroscope biases (\(b_{gN},b_{gD}\)), a \black slightly \black different behavior was 
\black noticed though. As depicted in Figs. \ref{fig:estbgN2} and \ref{fig:estbgD2}, only \black 
TRIAD-CBE \black and OPT-TRIAD-CBE \black 
provided estimates that converged to a \black steady state \black value within the 300-second interval. The other methods, \black in turn, \black produced estimates that showed only a tendency to converge toward the values estimated by TRIAD-CBE \black and OPT-TRIAD-CBE (notice the difference in the scales of each subplot of Figs. \ref{fig:estbgN2} and \ref{fig:estbgD2}). \black
This aspect, although less pronounced, was observed in both the simulations and the previous experiment (Figs. \ref{fig:align_bias_sim} and \ref{fig:misalign}) and can be attributed to the lower ``estimability'' of these gyroscope bias components compared to the accelerometer bias (\(b_{aD}\))\black\cite{silva17}. 

\begin{figure}[t!]
    \centering
    \includegraphics[width=\columnwidth]{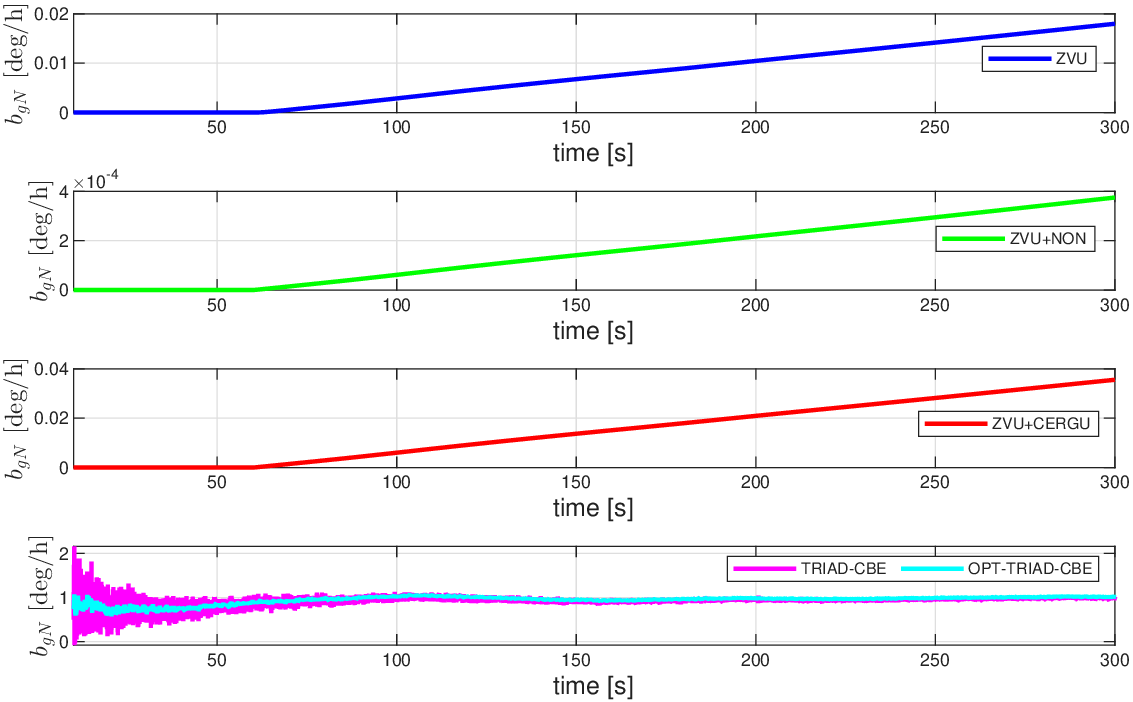} 
    \caption{
    \black North gyroscope bias estimates 
    (modified ASV experiment).}
    \label{fig:estbgN3} 
\end{figure}

\begin{figure}[t!]
    \centering
    \includegraphics[width=\columnwidth]{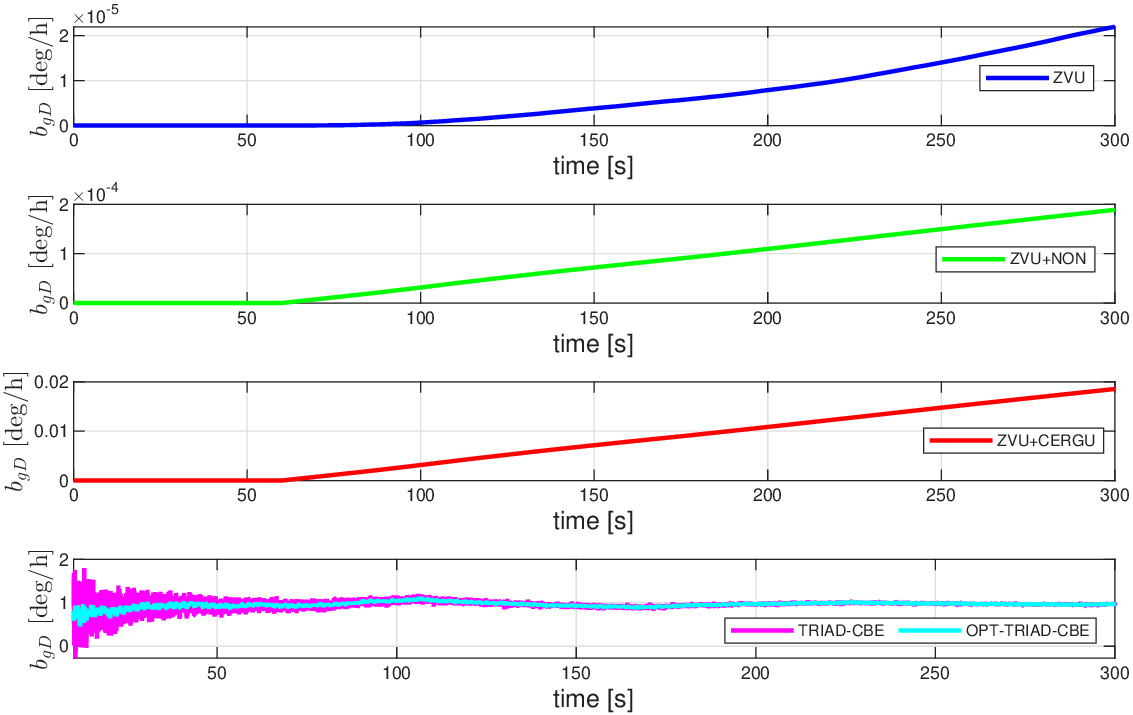} 
    \caption{
    \black Down gyroscope bias estimates 
    (modified ASV experiment).}
    \label{fig:estbgD3} 
\end{figure}

\black In fact, \black several factors may have contributed to the observed performance difference between the \black CA and FA methods \black 
in estimating the gyroscope biases. In addition to the relatively short simulation time (300 s), the \black Primus \black gyroscopes \black of the ASV test were known to: \black 
(a) exhibit a \black noise level 
significantly higher (SD of approximately 600 deg/hour) than the one predicted by the manufacturer (Table \ref{tab:Primus300_spec}); \black as well as \black 
(b) 
very reduced residual biases (0.02 deg/h), which were \black 
partially compensated \black for \black during manufacturing. 
\black The influence of such an increased sensor noise level on the estimability and convergence rate of EKF-based FA states has already been discussed by Frutuoso \textit{et al.} in a recent contribution in the area of AUVs \cite{frutuoso25}, and corroborates the verifications outlined here. 
\black With respect to the reduced magnitudes of the residual gyroscope biases, as the latter 
may have impaired our fair assessment of the performance of the estimation methods, we decided to rerun the ASV test by purposely 
corrupting the gyroscope signals with additional biases of 1 deg/h. The estimates of these biases, obtained by the evaluated methods, are presented in Figs. \ref{fig:estbgN3} and \ref{fig:estbgD3}, which 
clearly show 
that only 
TRIAD-CBE and OPT-TRIAD-CBE 
were able to provide us with estimates that converged to the 
purposely corrupted values of the biases.

\begin{figure}[t!]
    \centering
    \includegraphics[trim=1.2cm 0cm 1.5cm 0cm, clip, width=\columnwidth]{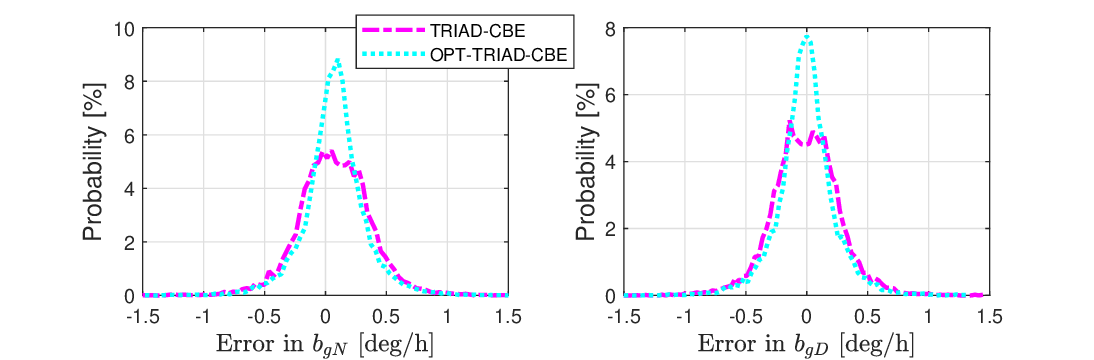} 
    \caption{
    \black Gyroscope bias error distributions 
    (modified MC \black simulation).}
    \label{fig:estbiasMC}
\end{figure}

Regarding the estimates of the gyroscope biases obtained by TRIAD-CBE and OPT-TRIAD-CBE, another aspect must be highlighted (which has not been observed in the previous simulations and real-world experiment). As indicated by the smaller dispersions, i.e., SDs, of Figs. \ref{fig:estbgN2}-\ref{fig:estbgD3}, for short alignment time intervals (less than 50 seconds) OPT-TRIAD-CBE 
provided us with more precise 
estimates than TRIAD-CBE. 
Aiming at replicating/explaining such behavior, 
a new 
MC simulation was performed, which adopted three main modifications w.r.t. to the original MC simulation of Section \ref{sec:res}: (a) a shorter alignment time (30 s); (b) gyroscope signals corrupted by increased levels of noise (white noises with SDs of 600 deg/h); and (c) gyroscope signals additionally corrupted by time-correlated noises (after a careful inspection/characterization of Primus stochastic errors, we identified the existence of the latter in the outputs of the gyroscopes). 
In this new MC simulation, an auto-regressive model \cite{nassar2003modeling} was used to represent these time-correlated error components, and 
the estimated 
PDFs of the gyroscope biases are given in Fig. \ref{fig:estbiasMC}. As can be observed,  the estimates provided by 
OPT-TRIAD-CBE 
were more precise (in agreement with the optimized nature of this method, compared to ordinary TRIAD-CBE), corroborating the results obtained in the modified ASV test.

Briefly, \black the results of the ASV experiment indicate that all 
methods \black under investigation \black 
performed comparably in \black determining \black the INS initial attitude as well as the \black estimable, i.e., down, \black 
component of the accelerometer biases. 
Due primarily to the error characteristics of the gyroscopes \black though, only TRIAD-CBE and specially OPT-TRIAD-CBE were \black 
able to provide \black us with sufficiently accurate and stable \black north and down bias estimates within the 300-second interval. \black Interestingly, no significant differences were noticed between TRIAD-CBE and OPT-TRIAD-CBE across all simulations and real-world experiments\black \black, except when a reduced alignment time was considered (30 seconds), as well as the existence of time-correlated noise in the outputs on the inertial sensors, 
a condition wherein 
OPT-TRIAD-CBE 
proved to perform better (higher precision), which is in line with the optimized nature of such method.


\black \section{Conclusion} \label{sec:con}

In this paper, \black we proposed \black an optimized version of 
TRIAD-CBE CA, 
\black as well as \black a new 
approach to \black the \black EKF-based FA \black 
of \black INSs with application to quasi-static AUVs/ASVs. \black 
The \black 
\black approaches were \black based on the \black optimized use/\black measurement augmentation of \black 
the \black NON \black 
errors \black originated from \black ordinary \black 
TRIAD-
derived \black 
rotation \black matrix, \black which are directly correlated to some \black 
bias components of the inertial sensors. As \black verified from both simulated (including a comprehensive MC analysis) and experimental tests \black (employing both a high-end turntable and a real-world ASV sitting on a pier), \black the proposed methods (referred to as \black OPT-TRIAD-CBE and \black ZVU-NON) \black 
showed similar performance to the 
\black baseline methods (TRIAD-CBE, \black ZVU and ZVU+CERGU), at least in terms of misalignment accuracy and precision. 

As for the convergence rate performance, the proposed 
approaches significantly outperformed the benchmarking methods, speeding up the estimation of the down misalignment and the 
\black ``estimable" \black inertial sensor biases \black 
from minutes 
to just a few seconds. \black Despite the effectiveness (and admitted optimality) of the proposed FA method, the conducted MC simulation also evidenced that 
\black OPT-TRIAD-CBE (as well as the non-optimal TRIAD-CBE) \black performed even 
better than \black the \black ZVU+NON-based FA, both in terms of accuracy/precision (for the estimation of the inertial sensor biases) and convergence rate (for the estimation of the misalignments and inertial sensor biases). Such verifications have no precedent in the previous literature, and stress the practical relevance of \black both TRIAD-CBE and OPT-TRIAD-CBE \black for the purpose of 
INS alignment and calibration \black of quasi-static AUVs/ASVs. \black 

\black With respect to the methods' precision, we were able to justify why the proposed ZVU+NON-based FA performed worse than TRIAD-CBE \black and OPT-TRIAD-CBE, \black as the former is function of noisy terms with increased uncertainties (high-order derivatives of the ``estimable" velocity errors). \black As a direction for future research, we \black ought to \black 
evaluate whether the worst performance of ZVU+NON FA w.r.t. \black OPT-TRIAD-CBE \black CA, in terms of convergence rate and bias estimation accuracy \black(see the biased MC distributions of Fig. \ref{fig:align_bias_mc}, for instance), is associated \black 
(or not) to the correlation \black that ends up existing between \black process and measurement noise (which violates \black basic \black 
assumptions 
\black upon which the conventional EKF is built). 
The extension/adaptation of the proposed OPT-TRIAD-CBE for the alignment problem of in-motion AUVs/ASVs is another topic worthy of investigation.

\bibliographystyle{IEEEtran}
\bibliography{IEEEabrv,AUV}

@book{S2000,
	title =     {Strapdown Analytics},
	author =    {Paul G. Savage},
	publisher = {Strapdown Associates},
	isbn =      {},
	year =      {2000},
	series =    {},
}

@article{silva16,
  author = {F. O. Silva and E. M. Hemerly and W. C. Leite\;Filho},
  title = {Error Analysis of Analytical Coarse Alignment Formulation for Stationary {SINS}},
  journal = {IEEE T. Aero. Elec. Sys.},
  volume = {52},
  number = {4},
  pages = {1777-1796},
  month = {Aug.},
  year = {2016}
}

@book{jekeli00,
  author = {C. Jekeli},
  title = {Inertial Navigation Systems with Geodetic Applications},
  publisher = {Walter de Gruyter GmbH \& Co.},
  year = {2000}
}

@book{markley14,
  author = {F. L. Markley and J. L. Crassidis},
  title = {Fundamentals of Spacecraft Attitude Determination and Control},
  publisher = {Springer},
  address = {New York},
  year = {2014},
  chapter = {5}
}

@article{silva18TIM,
  author = {F. O. Silva and E. M. Hemerly and W. C. Leite\;Filho and H. K. Kuga},
  title = {A Fast In-Field Coarse Alignment and Bias Estimation Method for Stationary Intermediate-Grade {IMUs}},
  journal = {IEEE T. Instrum. Meas.},
  volume = {67},
  number = {4},
  pages = {831-838},
  year = {2018}
}

@article{bar-itzhack88,
  author = {I. Y. Bar-Itzhack and N. Berman},
  title = {Control Theoretic Approach to Inertial Navigation Systems},
  journal = {J. Guid. Control},
  volume = {11},
  number = {3},
  pages = {237-245},
  year = {1988}
}

@book{groves13,
  author = {P. D. Groves},
  title = {Principles of {GNSS}, Inertial, and Multisensor Integrated Navigation Systems},
  publisher = {Artech House Remote Sens. Libr.},
  address = {London},
  year = {2013}
}

@article{silva17,
  author = {F. O. Silva and E. M. Hemerly and W. C. Leite\;Filho},
  title = {On the Error State Selection for Stationary {SINS} Alignment and Calibration {Kalman} Filters— Part {II}: Observability/Estimability Analysis},
  journal = {Sensors},
  volume = {17},
  number = {439},
  pages = {1-34},
  year = {2017}
}

@book{titterton04,
  author = {D. H. Titterton and J. L. Weston},
  title = {Strapdown Inertial Navigation Technology},
  publisher = {Inst. Electr. Eng.},
  address = {Reston},
  year = {2004}
}

@article{silva18MEAS,
  author = {F. O. Silva and E. M. Hemerly and W. C. Leite\;Filho},
  title = {On the Measurement Selection for Stationary {SINS} Alignment {Kalman} Filters},
  journal = {Measurement},
  volume = {130},
  pages = {82-93},
  year = {2018}
}

@inproceedings{silva14,
  author = {F. O. Silva and E. M. Hemerly and W. C. Leite\;Filho and R. A. J. Chagas},
  title = {An Improved Stationary Fine Self-Alignment Approach for {SINS} Using Measurement Augmentation},
  booktitle = {Proc. Braz. Conf. Autom. ({SBA})},
  address = {Belo Horizonte},
  year = {2014},
  pages = {790-795}
}

@book{brown12,
  author = {R. G. Brown and P. Y. C. Hwang},
  title = {Introduction to Random Signals and Applied {Kalman} Filtering},
  publisher = {John Wiley \& Sons, Inc.},
  year = {2012}
}

@article{silva25,
  author = {F. O. Silva and A. H. A. Maia and J. B. Uwineza and F. S. Rahman and Z. Jiang and W. Hu and J. A. Farrell},
  title = {Dual-Antenna {GNSS}-Aided {INS} Stationary Alignment with Sensor Parameter Estimation},
  journal = {IEEE T. Instrum. Meas.},
  volume = {74},
  number = {8501716},
  pages = {1-16},
  year = {2025}
}

@article{jiang1998,
  author  = {Y.F. Jiang},
  title   = {Error Analysis of Analytic Coarse Alignment Methods},
  journal = {IEEE T. Aero. Elec. Sys.},
  volume  = {34},
  number  = {1},
  pages   = {334-337},
  month   = {Jan.},
  year    = {1998},
  doi     = {10.1109/7.640292}
}

@article{SILVA201745,
title = {{On the error state selection for stationary SINS alignment and calibration Kalman filters – part I: Estimation algorithms}},
journal = {Aerosp. Sci. Technol.},
volume = {61},
pages = {45-56},
year = {2017},
issn = {1270-9638},
doi = {https://doi.org/10.1016/j.ast.2016.11.019},
author = {Felipe O. Silva and Elder M. Hemerly and Waldemar C. {Leite\;Filho}},
}

@article{engelsman2023information,
  title={{Information-aided inertial navigation: A review}},
  author={Engelsman, Daniel and Klein, Itzik},
  journal = {IEEE T. Instrum. Meas.},
  volume={72},
  pages={1-18},
  year={2023},
  publisher={IEEE}
}

@article{RefB6,
title = {{Rotary INS self-alignment method based on backtracking filtering under large misalignment angle}},
author = {Chenming Zhang and Jie Li and Kaiqiang Feng and Xiaokai Wei},
journal = {Measurement},
volume = {231},
pages = {114537},
year = {2024},
}

@article{RefB7,
title = {{Analytical quaternion-based bias estimation algorithm for fast and accurate stationary gyro-compassing}},
author = {Mohammadkarimi, H and Mozafari, S and Alizadeh, M H},
journal = {Scientific Reports},
volume = {14},
pages = {2045-2322},
year = {2024},
}

@article{RefB8,
title = {{A high-accuracy initial alignment method based on backtracking process for strapdown inertial navigation system}},
journal = {Measurement},
volume = {201},
pages = {111712},
year = {2022},
author = {Yusen Lin and Lingjuan Miao and Zhiqiang Zhou},
}

@ARTICLE{RefB9,
  author={Klein, Itzik and Bar-Shalom, Yaakov},
  journal={IEEE Access}, 
  title={{INS} Fine Alignment With Low-Cost Gyroscopes: Adaptive Filters for Different Measurement Types}, 
  year={2021},
  volume={9},
  number={},
  pages={79021-79032},
}

@ARTICLE{paull14,
  author={Paull, Liam and Saeedi, Sajad and Seto, Mae and Li, Howard},
  journal={IEEE J. Ocean. Eng.}, 
  title={{AUV} Navigation and Localization: A Review}, 
  year={2014},
  volume={39},
  number={1},
  pages={131-149},
  doi={10.1109/JOE.2013.2278891}}

@ARTICLE{pinto22,
  author={Pinto, Marc A. and Verrier, Laurent},
  journal={IEEE J. Ocean. Eng.}, 
  title={Interferometric {D}oppler Velocity Sonar for Low Bias Long Range Estimation of Speed Over Seabed}, 
  year={2022},
  volume={47},
  number={3},
  pages={767-779},
  doi={10.1109/JOE.2021.3130662}}

@ARTICLE{meurer20,
  author={Meurer, Christian and Francisco Fuentes-Pérez, Juan and Palomeras, Narcís and Carreras, Marc and Kruusmaa, Maarja},
  journal={IEEE J. Ocean. Eng.}, 
  title={Differential Pressure Sensor Speedometer for Autonomous Underwater Vehicle Velocity Estimation}, 
  year={2020},
  volume={45},
  number={3},
  pages={946-978},
  doi={10.1109/JOE.2019.2907822}}

@ARTICLE{fallon13,
  author={Fallon, Maurice F. and Folkesson, John and McClelland, Hunter and Leonard, John J.},
  journal={IEEE J. Ocean. Eng.}, 
  title={Relocating Underwater Features Autonomously Using Sonar-Based {SLAM}}, 
  year={2013},
  volume={38},
  number={3},
  pages={500-513},
  doi={10.1109/JOE.2012.2235664}}

@ARTICLE{song18,
  author={Song, Zhuoyuan and Mohseni, Kamran},
  journal={IEEE J. Ocean. Eng.}, 
  title={Long-Term Inertial Navigation Aided by Dynamics of Flow Field Features}, 
  year={2018},
  volume={43},
  number={4},
  pages={940-954},
  doi={10.1109/JOE.2017.2766900}}

@ARTICLE{mcphail09,
  author={McPhail, Stephen D. and Pebody, Miles},
  journal={IEEE J. Ocean. Eng.}, 
  title={Range-Only Positioning of a Deep-Diving Autonomous Underwater Vehicle From a Surface Ship}, 
  year={2009},
  volume={34},
  number={4},
  pages={669-677},
  doi={10.1109/JOE.2009.2030223}}

@ARTICLE{sarda19,
  author={Sarda, Edoardo I. and Dhanak, Manhar R.},
  journal={IEEE J. Ocean. Eng.}, 
  title={Launch and Recovery of an Autonomous Underwater Vehicle From a Station-Keeping Unmanned Surface Vehicle}, 
  year={2019},
  volume={44},
  number={2},
  pages={290-299},
  doi={10.1109/JOE.2018.2867988}}

@inproceedings{nassar2003modeling,
  author    = {Nassar, S. and Schwarz, K. P. and Noureldin, A. and El-Sheimy, N.},
  title     = {Modeling Inertial Sensor Errors Using Autoregressive ({AR}) Models},
  booktitle = {Proceedings of the Institute of Navigation National Technical Meeting (ION NTM)},
  year      = {2003},
  address   = {Anaheim, CA},
}

@article{frutuoso25,
author = {Frutuoso, Adriano and Silva, Felipe O. and de Barros, Ettore A.},
title = {Assessment of Maneuvering Influence on the Fine Alignment of Autonomous Underwater Vehicle},
journal = {J. Field Robot.},
year = {2025},
volume = {n/a},
number = {22551},
pages = {1-25},
doi = {https://doi.org/10.1002/rob.22551},
url = {https://onlinelibrary.wiley.com/doi/abs/10.1002/rob.22551},
eprint = {https://onlinelibrary.wiley.com/doi/pdf/10.1002/rob.22551}
}

@article{ZHANG2023113861,
title = {Autonomous Underwater Vehicle navigation: A review},
journal = {Ocean Engineering},
volume = {273},
pages = {113861},
year = {2023},
issn = {0029-8018},
doi = {https://doi.org/10.1016/j.oceaneng.2023.113861},
url = {https://www.sciencedirect.com/science/article/pii/S0029801823002457},
author = {Bingbing Zhang and Daxiong Ji and Shuo Liu and Xinke Zhu and Wen Xu}
}

@book{golub2013matrix,
  title={Matrix Computations},
  author={Golub, Gene H. and Van Loan, Charles F.},
  edition={4},
  year={2013},
  publisher={Johns Hopkins University Press}
}

@ARTICLE{baram88,
  author={Baram, Y. and Kailath, T.},
  journal={IEEE Trans. Autom. Control}, 
  title={Estimability and regulability of linear systems}, 
  year={1988},
  volume={33},
  number={12},
  pages={1116-1121},
  doi={10.1109/9.14433}}

@ARTICLE{goshen-meskin92,
  author={Goshen-Meskin, D. and Bar-Itzhack, I.Y.},
  journal={IEEE Trans. Autom. Control}, 
  title={On the connection between estimability and observability}, 
  year={1992},
  volume={37},
  number={8},
  pages={1225-1226},
  doi={10.1109/9.151112}}

@article{chernoff1952measure,
  title={A measure of asymptotic efficiency for tests of a hypothesis based on the sum of observations},
  author={Chernoff, Herman},
  journal={Ann. Math. Stat.},
  volume={23},
  number={4},
  pages={493--507},
  year={1952},
  publisher={JSTOR}
}


\begin{IEEEbiography}
[{\includegraphics[
width=1.22 in,
height=1.22 in,
clip,
keepaspectratio]{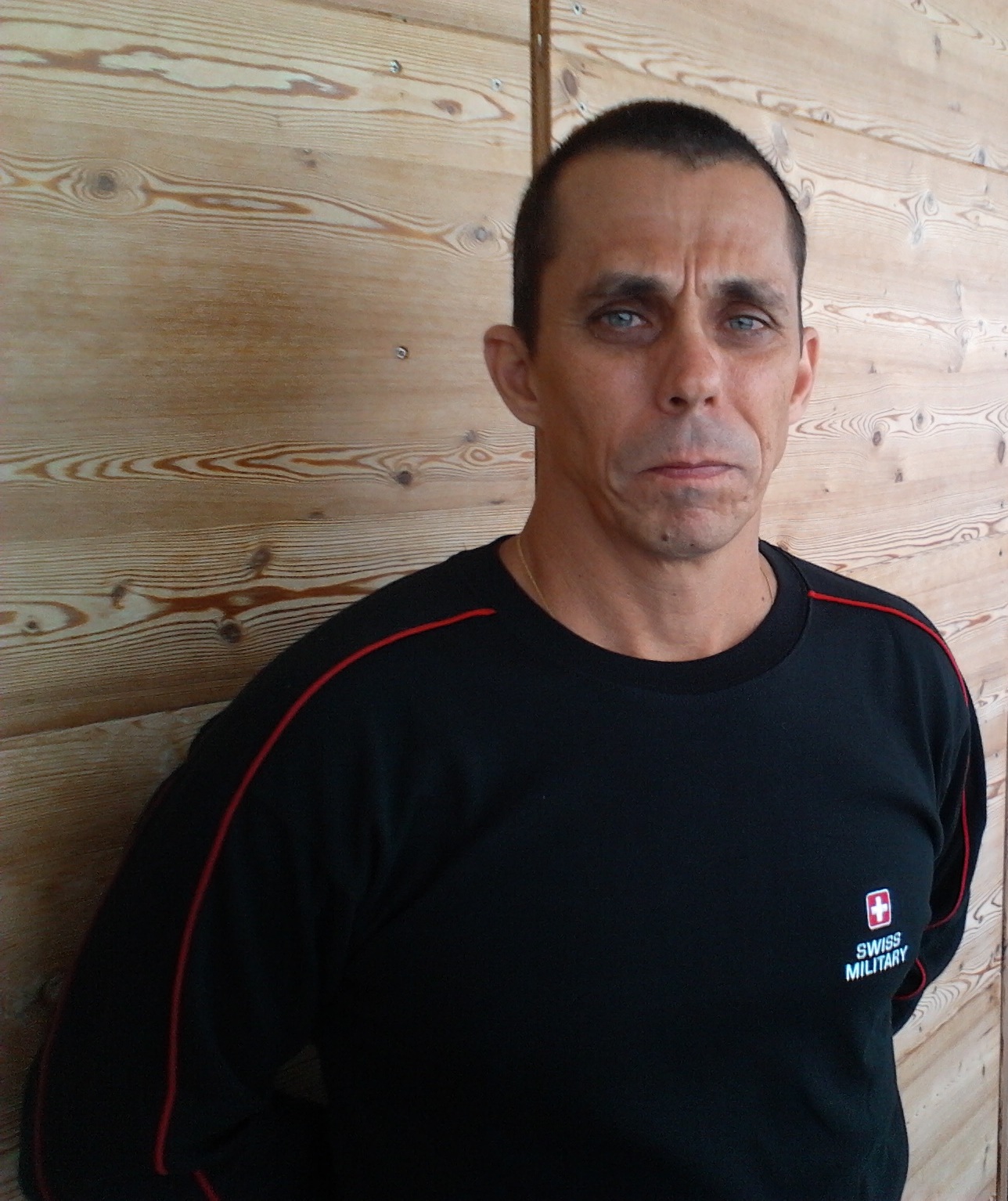}}]
{}

Carlos Renato Caputo Durão received the B.Sc. degree in Electronic Engineering from Universidade Gama Filho, Rio de Janeiro, Brazil, in 1985, the M.Sc. degree in Electrical Engineering from the Federal University of Rio de Janeiro (UFRJ), Brazil, in 1992, and the Ph.D. degree in Electrical Engineering from the same institution in 2009. He worked as a Researcher at the Navy Research Institute (IPqM), Brazil, from 1987 to 2025. He is currently a Collaborating Professor at the Military Institute of Engineering (IME), Rio de Janeiro, Brazil, and a Postdoctoral Researcher with the Graduate Program in Systems Engineering and Automation (PPGESISA), Federal University of Lavras (UFLA), Brazil. His research interests include guidance, navigation and control systems for autonomous vehicles, optimization, mathematical modeling, computer simulation, statistical inference, data analysis, and stochastic processes.

\end{IEEEbiography}

\begin{IEEEbiography}
[{\includegraphics[
width=1.22 in,
height=1.22 in,
clip,
keepaspectratio]{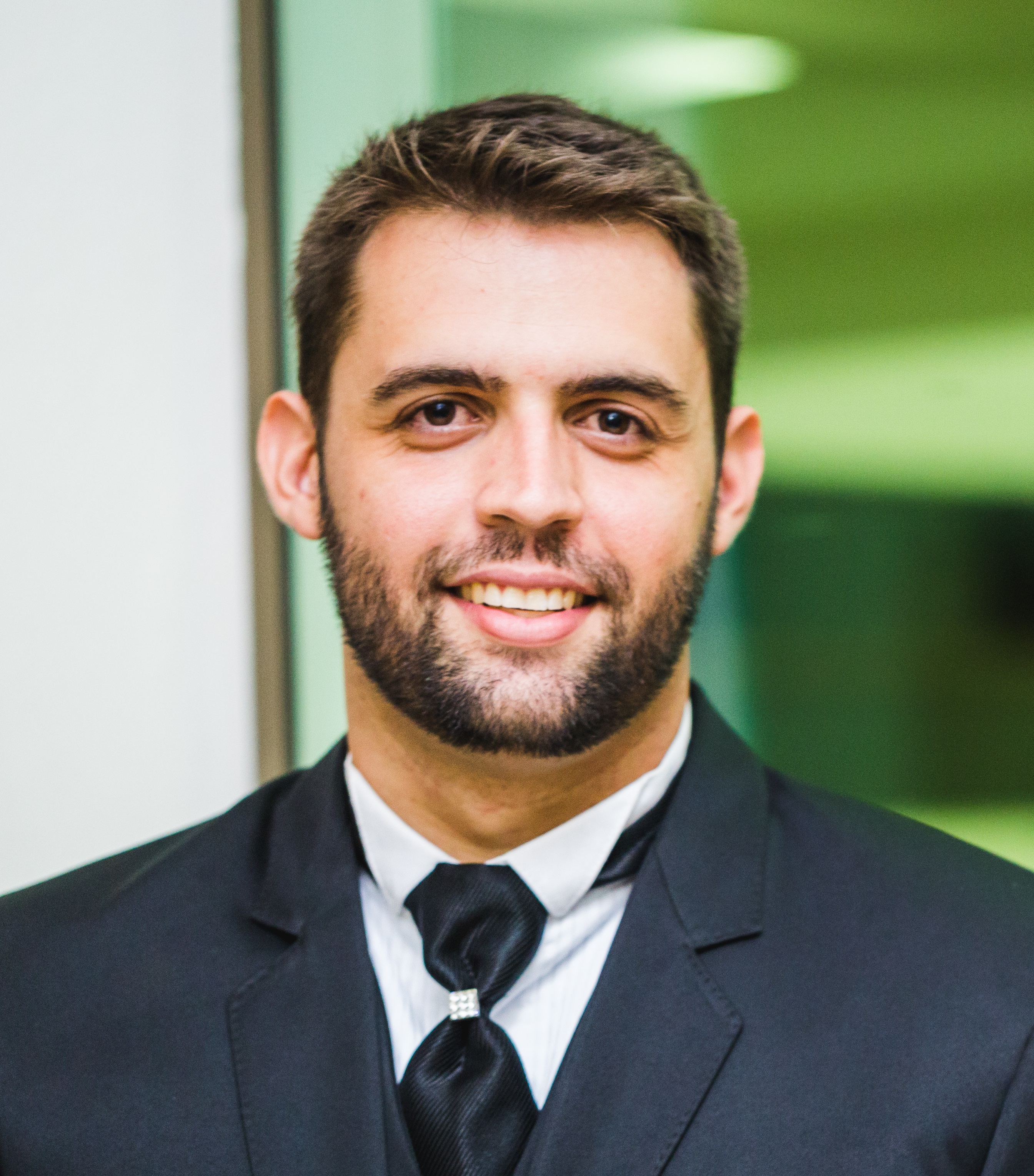}}]
{}

Felipe O. Silva received the B.S. degree (with honors) in Automatic Control Engineering from the Federal University of Itajubá, Brazil, in 2012; the M.S. degree in Systems Engineering from the National Institute of Applied Sciences, Center Val de Loire, France, in 2013; and the Ph.D. degree in Aeronautical and Mechanical Engineering from the Aeronautics Institute of Technology, Brazil, in 2016. In 2012, he was an intern with Alcatel-Lucent Bell Labs France, and from 2013 to 2014, an assistant researcher with the Institute of Aeronautics and Space, Brazil. Since 2014, he has been a Professor (currently Associate) with the Federal University of Lavras (UFLA), Brazil. He held visiting professorships at the Central School of Nantes, France, in 2018, and the University of California, Riverside, USA, from 2018 to 2019. Since 2020, he has been a CNPq productivity fellow. In 2025, he became the coordinator of the Graduate Program in Systems Engineering and Automation at UFLA, and a Member of the Advisory Board of the Technology Innovation Institute's Autonomous Robotics Research Center in Abu Dhabi, UAE. His research interests include state estimation, stochastic filtering, sensor fusion, instrumentation, and robotics, with applications to GNC systems, INS, GNSS, connected autonomous vehicles, and precision agriculture. Dr. Silva is a Senior Member of the IEEE and a member of ION, ISIF, IFAC, SBA, and SBC, as well as a founding member of the Brazilian Robotics Society.

\end{IEEEbiography}

\begin{IEEEbiography}
[{\includegraphics[
width=1.10 in,
height=1.40 in,
clip,
keepaspectratio]{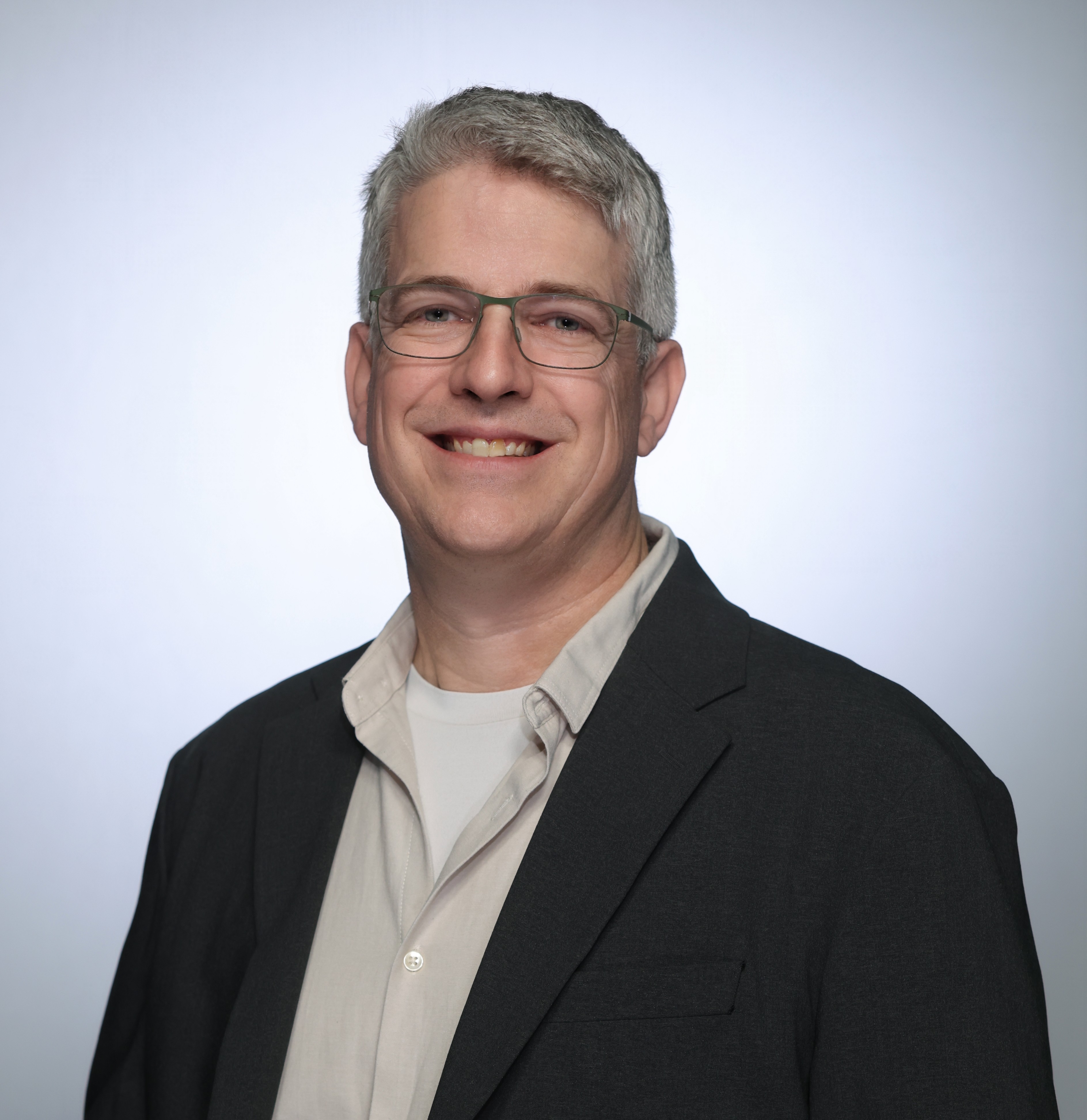}}]
{}

Itzik Klein received the B.Sc. and M.Sc. degrees in Aerospace Engineering from the Technion - Israel Institute of Technology, Haifa, Israel, in 2004 and 2007, respectively, and a Ph.D. degree in Geo-information Engineering from the Technion - Israel Institute of Technology, in 2011. He is an Associate Professor at the University of Haifa, where he heads the Autonomous Navigation and Sensor Fusion Lab (ANSFL) and serves as Chair of the Hatter Department of Marine Technologies in the Charney School of Marine Sciences. He also directs the University’s AI Research Center. He is an IEEE Senior Member and an IEEE Oceanic Engineering Society (OES) Distinguished Lecturer. He serves on the editorial boards of the IEEE Transactions on Instrumentation and Measurement (TIM), IEEE Journal of Indoor and Seamless Positioning and Navigation (J-ISPIN) and Elsevier Results in Engineering, and is the founding editor in the area of sensing and perception for automotive applications for the IEEE Open Journal of Vehicular Technology editorial board.  Prior to joining the University of Haifa, he spent more than 15 years in leading Israeli technology companies, specializing in navigation systems. His work bridges industry and academia, with deep expertise in inertial sensing, sensor fusion, robotics, and the integration of artificial intelligence into advanced navigation systems.

\end{IEEEbiography}

\begin{IEEEbiography}
[{\includegraphics[
width=1.22 in,
height=1.22 in,
clip,
keepaspectratio]{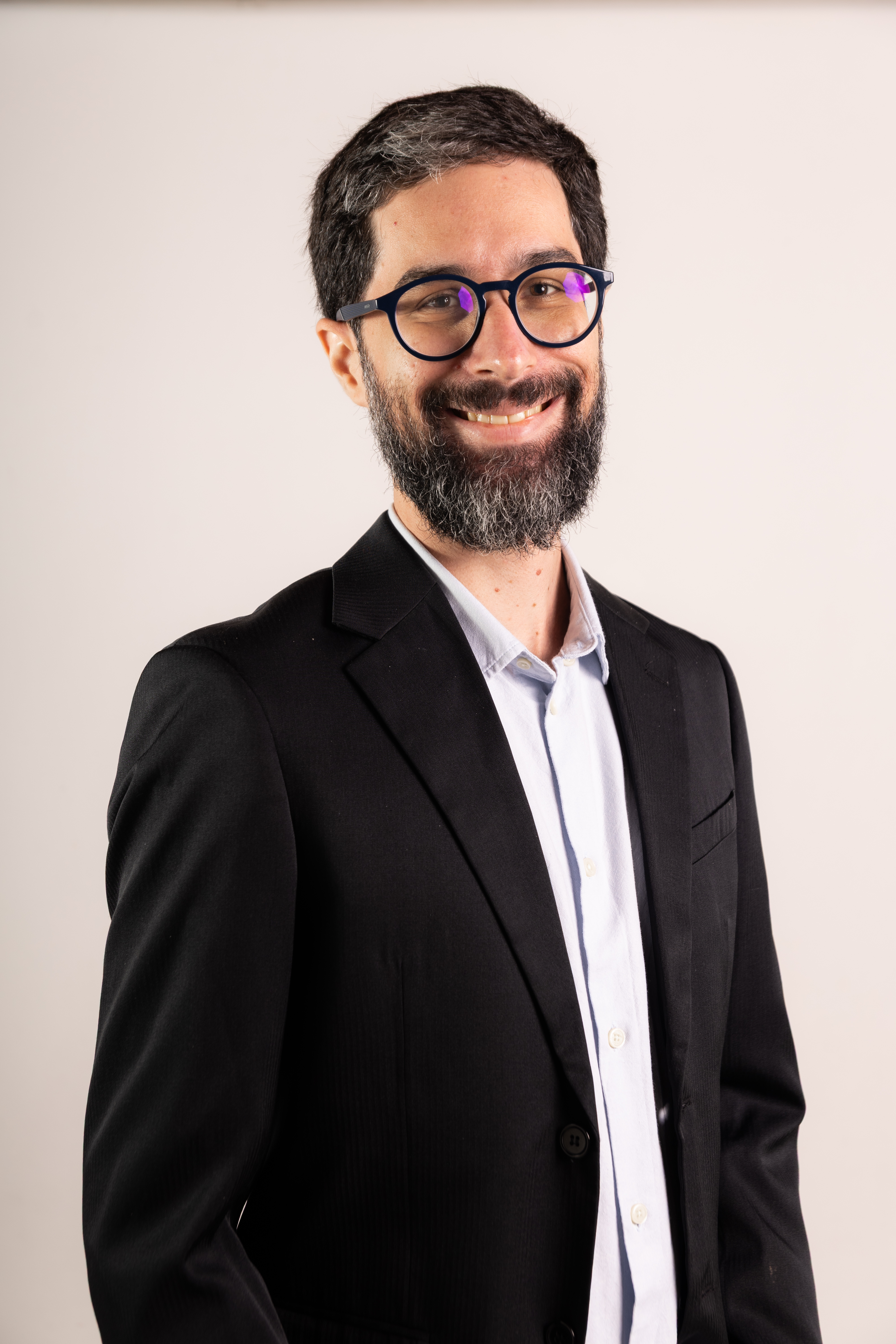}}]
{}

Vinícius M.G.B. Cavalcanti received his BS degree in Automatic Control Engineering from the Fluminense Federal Institute of Education, Science and Technology (IFF), Campos dos Goytacazes, Brazil, in 2015; MS degree in Electrical Engineering (Automation and Control line of research) and PhD in Defense Engineering (Mechatronics and Weapon Systems line of research) from the Military Institute of Engineering (IME), Rio de Janeiro, Brazil, in 2017 and 2021, respectively. In 2023, he was a Postdoctoral Researcher in the Department of Automatics at the Federal University of Lavras (UFLA), in Lavras, Brazil. Since 2023, he has been an Assistant Professor at the Fluminense Federal Institute of Education, Science and Technology (IFF), Itaboraí, Brazil. Since 2025, he has been a Postdoctoral Researcher in the Department of Electrical Engineering at the Londrina State University (UEL), Londrina, Brazil. His research interests include robust and linear parameter-varying control; stochastic filtering and error modeling; sensor fusion and guidance, navigation, and control (GNC) systems applications.

\end{IEEEbiography}

\begin{IEEEbiography}
[{\includegraphics[
width=1.22 in,
height=1.22 in,
clip,
keepaspectratio]{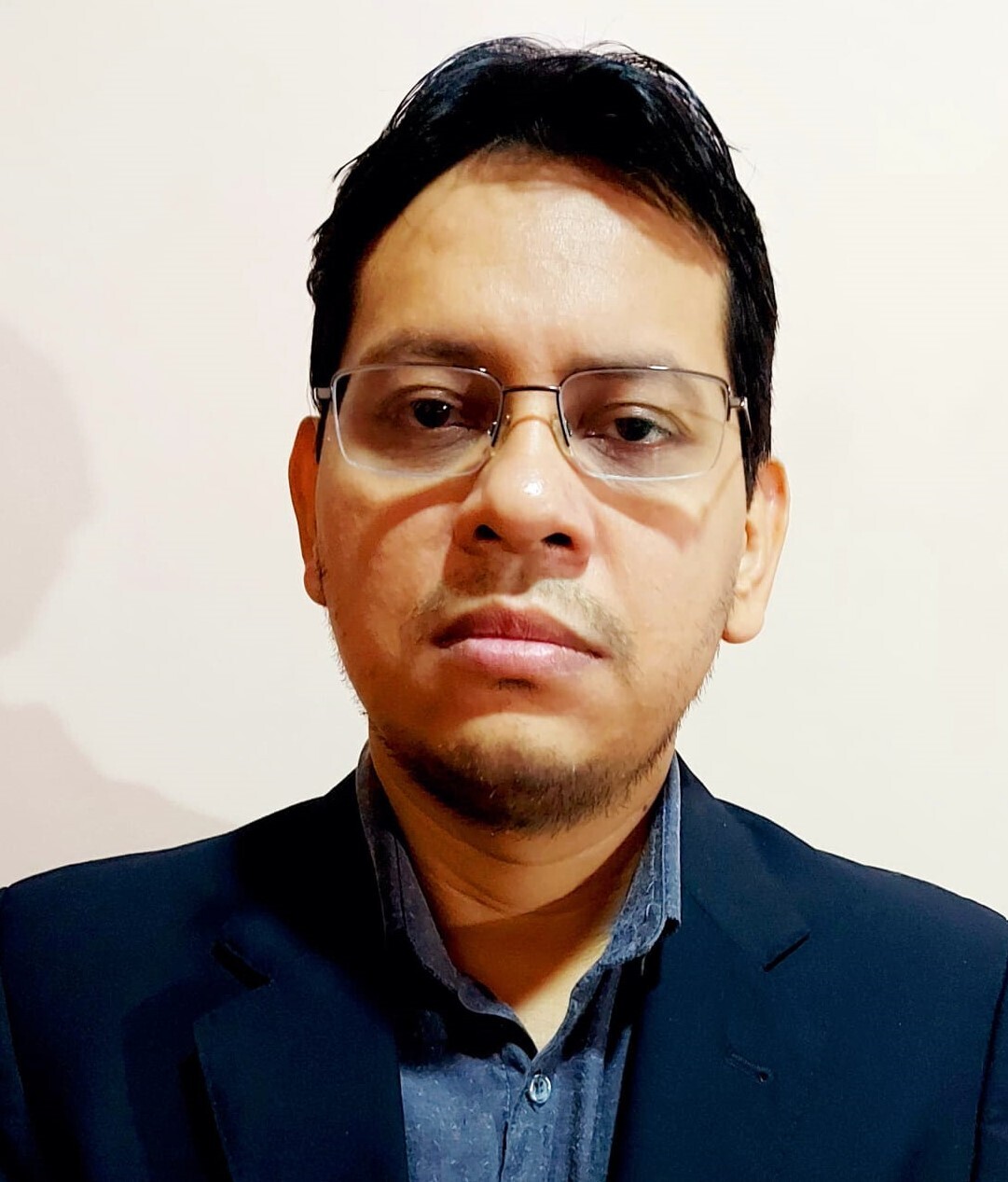}}]
{}

Adriano Frutuoso received the B.S. and M.S. degrees in electrical engineering from the Federal University of Amazonas (UFAM), Brazil, in 2012 and 2015, respectively, and the Ph.D degree in control and mechanical automation engineering from the University of São Paulo (USP), Brazil, in 2023. He is currently an Associate Professor with the Federal Institute of Education, Science and Technology of Amazonas, Manaus, Brazil. His research interests include autonomous navigation, alignment techniques of inertial navigation systems, and sensory fusion.

\end{IEEEbiography}

\begin{IEEEbiography}
[{\includegraphics[
width=1.10 in,
height=1.50 in,
clip,
keepaspectratio]{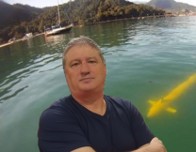}}]
{}

Ettore A. de Barros received the Graduate degree in engineering from the Department of Naval Architecture and Marine Engineering, University of São Paulo, São Paulo, Brazil, in 1985, and the Doctor of Engineering degree in naval architecture and ocean engineering from the University of Tokyo, Tokyo, Japan, in 1994. He was a Research Scientist with the Institute of Industrial Science, University of Tokyo, from 1995 to 1996, and held visiting positions with the Control System Division of Mitsubishi Heavy Industries, Nagasaki, Japan, from 1994 to 1995, and with the Dynamical Systems and Ocean Robotics Laboratory of the Institute for Systems and Robotics, Instituto Superior Tecnico, Lisbon, Portugal, from 2003 to 2004. Since 2000, he has been with the Mechatronics Engineering Department, University of São Paulo, São Paulo, Brazil, where he coordinates the Unmanned Vehicles Laboratory. His research specialty areas include dynamics and control and navigation of unmanned marine vehicles.

\end{IEEEbiography}

\begin{IEEEbiography}
[{\includegraphics[
width=1.10 in,
height=1.22 in,
clip,
keepaspectratio]{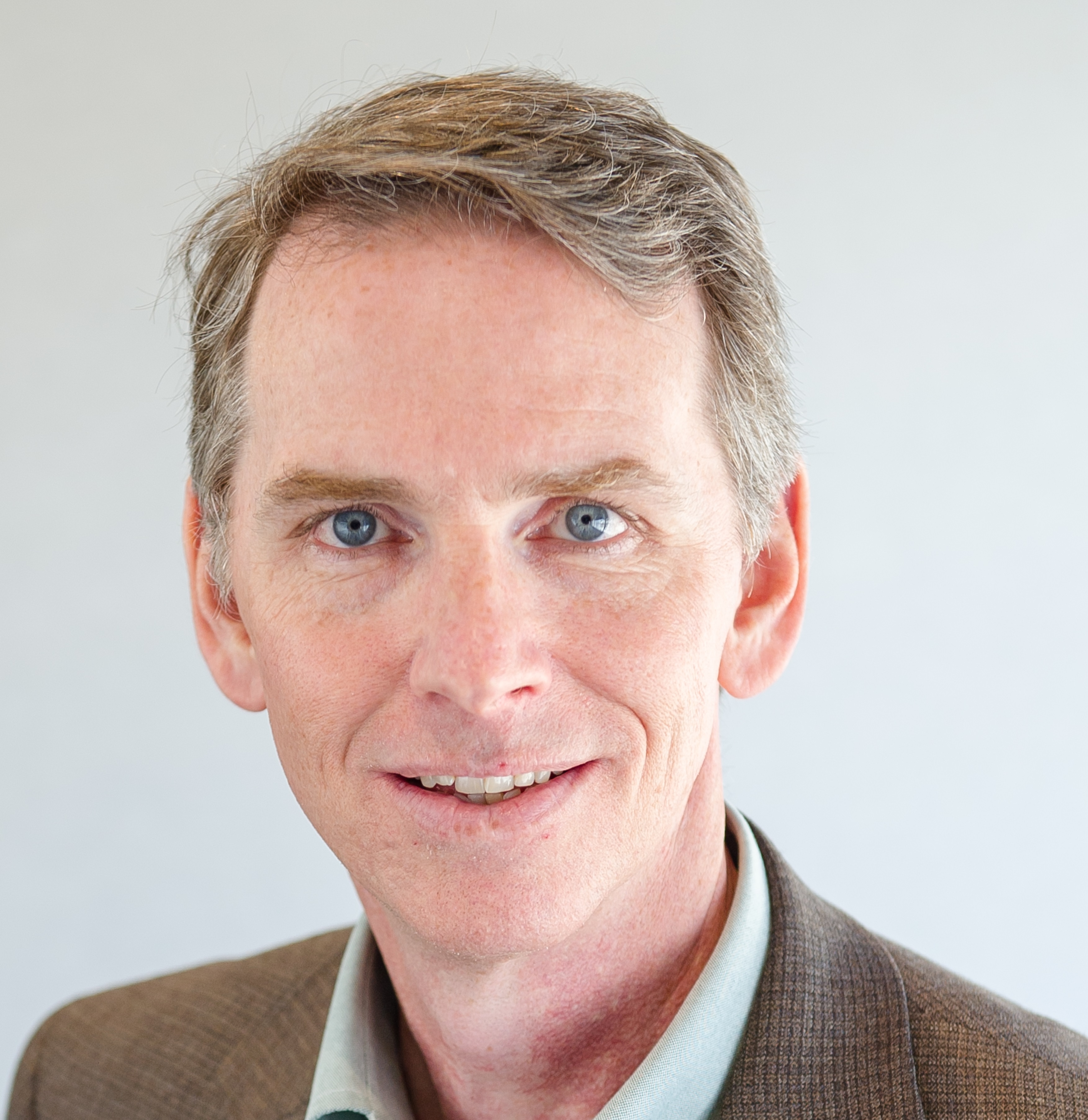}}]
{}

Jay A. Farrell received B.S. degrees in physics and electrical engineering from Iowa State University, and M.S. and Ph.D. degrees in electrical engineering from the University of Notre Dame. At Charles Stark Draper Lab, he was a principal engineer in the autonomous vehicles group, receiving the Engineering Vice President's Best Technical Publication Award in 1990, and Recognition Awards for Outstanding Performance and Achievement in 1991 and 1993. At the University of California, Riverside (1994-2024), he was the KA Endowed Professor in the Department of Electrical and Computer Engineering, served nine years as department chair, and four years as associate dean for academic personnel. He is now employed at Zoox, Inc. He served as General Chair of IEEE CDC 2012, President of IEEE CSS in 2014, and President of the American Automatic Control Council in 2020-2021. He is author of over 300 technical articles and three books. He was recognized as a GNSS Leader to Watch by GPS World Magazine, and is a Distinguished Member of IEEE CSS, a Fellow of the IEEE, a Fellow of AAAS, and a Fellow of IFAC.

\end{IEEEbiography}



\end{document}